%% file: neurips2026_arXiv_version.tex
\documentclass{article}

\PassOptionsToPackage{numbers,square}{natbib} % to enable nuumerical citation for rename style

 \usepackage[preprint]{neurips_2026}

\usepackage[utf8]{inputenc} % allow utf-8 input
\usepackage[T1]{fontenc}    % use 8-bit T1 fonts
\usepackage{hyperref}       % hyperlinks
\usepackage{url}            % simple URL typesetting
\usepackage{booktabs}       % professional-quality tables
\usepackage{amsfonts}       % blackboard math symbols
\usepackage{nicefrac}       % compact symbols for 1/2, etc.
\usepackage{microtype}      % microtypography
\usepackage[table]{xcolor}  % colors

\usepackage{amsmath,amssymb}
\usepackage{booktabs}   % \toprule, \midrule, \bottomrule, \cmidrule
\usepackage{graphicx}  % \resizebox
\usepackage{array}     % extended column formatting (good practice here)
\usepackage{caption}   % better control over captions
\usepackage{makecell}  % if you later tweak headers (not strictly needed here)
\usepackage{soul}

\usepackage{subcaption}
\usepackage{tikz}

\usepackage[most]{tcolorbox}
\usepackage[percent]{overpic}
\usepackage{url}            % or \usepackage{url} if you prefer \url instead of \path
\usepackage{hyperref}       % to use: \href{URL}{Name}

\graphicspath{{figures/illus/}}

\usepackage{pifont}            % To use right and x marks
\definecolor{lightgraybg}{gray}{0.96}
\definecolor{refcolor}{HTML}{0B6B72}
\definecolor{moderatecolor}{HTML}{454FAD}
\definecolor{highcolor}{HTML}{991C27}
\definecolor{softgreen}{rgb}{0.77, 0.84, 0.72}

\definecolor{lightgreenbg}{HTML}{D9FAD3}   % soft light-green background
\definecolor{darkgreenfg}{HTML}{0B6623}    % darker green for text (good contrast)
\newcommand{\highlightLR}[1]{%
  \colorbox{lightgreenbg}{\textcolor{darkgreenfg}{\textbf{#1}}}%
}

\title{REFORM-3D: A Representation-Centric Evaluation Framework for 3D Medical Vision Foundation Models}

\author{
  Maregu Assefa$^1$, Muzammal Naseer$^{1,2}$, Divya Velayudhan$^1$,\\
  \textbf{Gedamu Alemu Kumie}$^1$,
  \textbf{Naoufel Werghi}$^{1,}$\thanks{Corresponding author:
  \texttt{Naoufel.Werghi@ku.ac.ae}}\\
  $^1$Department of Computer Science, Khalifa University, Abu Dhabi, UAE\\
  $^2$University of Western Australia
}

\begin{document}

% original
% \maketitle

% modified to replace the * in the \thanks with \dagger:
\setcounter{footnote}{1}
\maketitle
\setcounter{footnote}{0}

\begin{abstract}
Evaluation of 3D medical foundation models is typically dominated by downstream task scores, which entangle representation quality with adaptation choices. This is especially critical in volumetric imaging, where spacing and anisotropy can substantially alter image statistics while the anatomy itself remains unchanged. We therefore introduce \textbf{REFORM-3D}, a \textbf{R}epresentation-centric \textbf{E}valuation \textbf{F}ramew\textbf{OR}k for benchmarking frozen 3D medical vision foundation \textbf{M}odels across three questions: Acquisition Robustness, asking whether features remain stable when spacing varies while anatomy is fixed; Cross-Modal Anatomical Alignment, asking whether labeled anatomical structures remain recoverable across modalities under balanced bidirectional retrieval; and Anatomical Transfer, asking whether organ identity remains recoverable from limited labeled support under case-disjoint evaluation. These are also pretraining questions, because what a 3D encoder learns depends on how volumes are geometrically presented during self-supervised pretraining. We further introduce Med3DINO encoder family, pretrained on $62.6$K unlabeled MR, CT, PET, CTA, and CBCT volumes under three explicit geometric assumptions: isotropic normalization, voxel-relative sampling, and spacing-aware resampling. Across Med3DINO and heterogeneous external encoders, REFORM-3D reveals module-specific strengths rather than a single model ordering. Within the controlled Med3DINO comparison, voxel-relative sampling yields lower embedding drift, whereas spacing-aware resampling favors anatomical transfer; semantic-transfer advantages depend on the dataset and feature type. Project link: \href{https://endarzboy.github.io/reform3d-website/}{\highlightLR{REFORM-3D}}.

\end{abstract}

\section{Introduction}
\label{sec:intro}
Evaluation of 3D medical vision foundation models is still dominated by downstream task scores~\cite{3dinov2,ayzenberg:dinov2_fewshot,li2025:meddinov3,perez2025:rad_dinov2}. Those scores are useful for showing whether a pretrained encoder can be adapted to a particular segmentation, classification, or retrieval pipeline, but they provide limited evidence about what the frozen representation itself encodes. End-task performance depends not only on pretrained features but also on decoder design, optimization choices, supervision level, and dataset split. As a result, a model can appear strong after adaptation without clarifying whether its features are organized by anatomy or rely on narrower cues that happen to be predictive under a particular evaluation protocol. 

\begin{figure}[!t]
    \centering
    \begin{minipage}[c]{0.52\textwidth}
        \centering

        \begin{tcolorbox}[
            colback=white,
            colframe=black,
            boxrule=0.4pt,
            arc=2pt,
            left=1pt,
            right=1pt,
            top=1pt,
            bottom=1pt,
            boxsep=0pt,
            enhanced,
            width=\linewidth
        ]
        \setlength{\tabcolsep}{0pt}
        \resizebox{\linewidth}{!}{%
            \begin{tabular}{@{}ccc@{}}
                {\scriptsize Reference (Isotropic)} &
                {\scriptsize Moderate anisotropy} &
                {\scriptsize Higher-anisotropy} \\

                \includegraphics[width=0.373\linewidth]{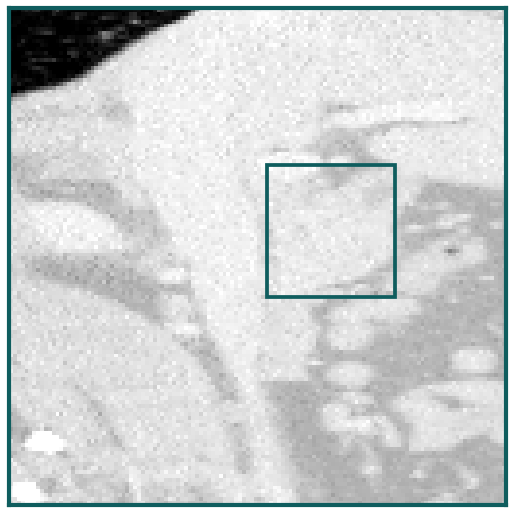} &
                \includegraphics[width=0.373\linewidth]{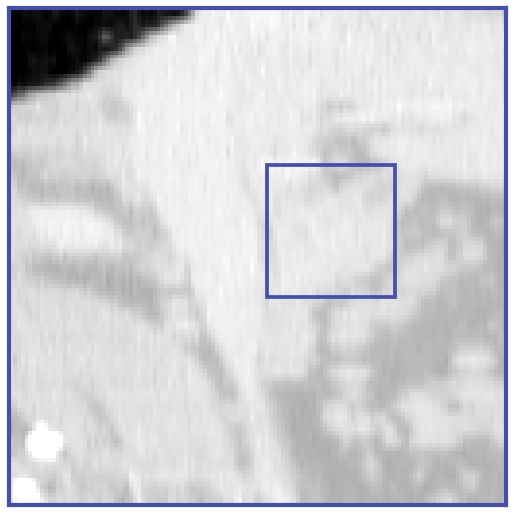} &
                \includegraphics[width=0.373\linewidth]{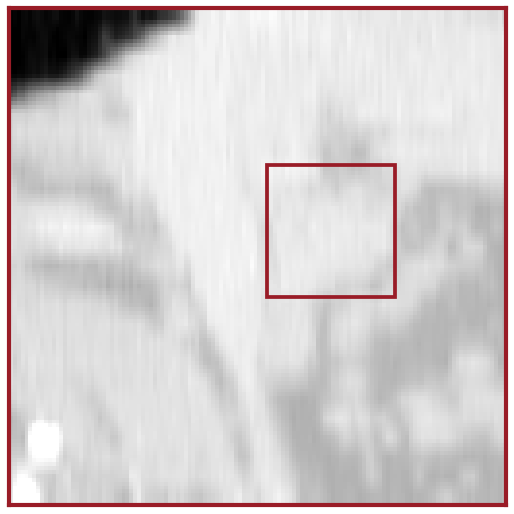} \\[-2pt]

                \begin{overpic}[width=0.373\linewidth]{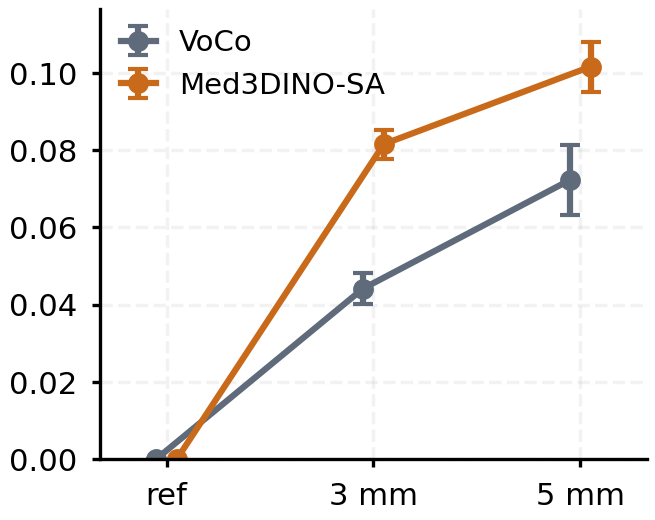}
                    \put(40,15){\scriptsize a. Cosine Drift}
                \end{overpic} &
                \begin{overpic}[width=0.373\linewidth]{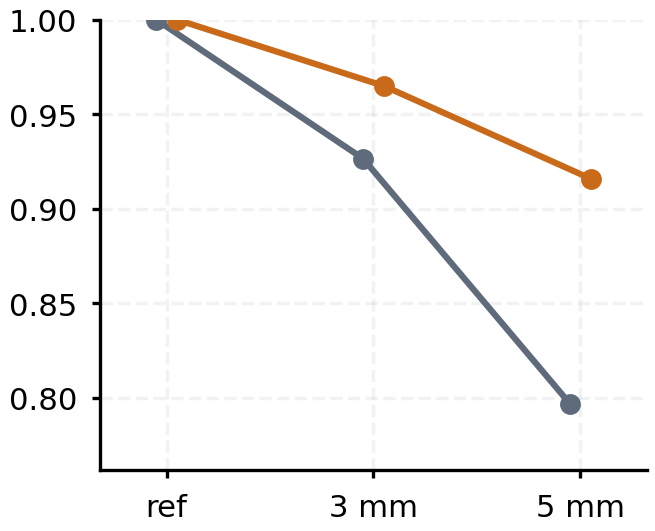}
                    \put(26,15){\scriptsize b. CKA}
                \end{overpic} &
                \begin{overpic}[width=0.373\linewidth]{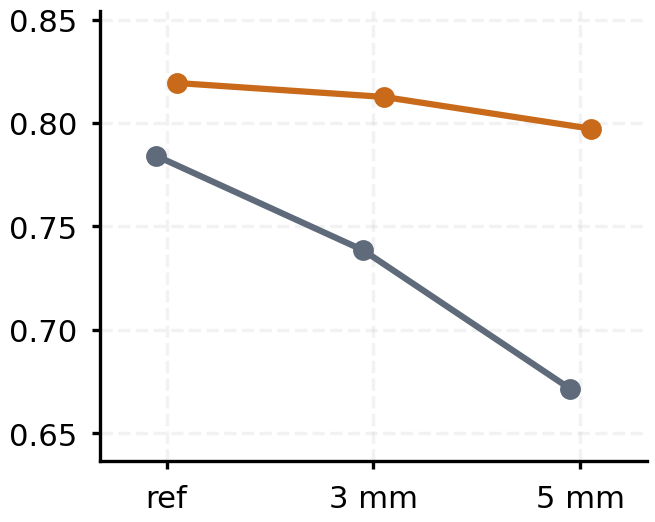}
                    \put(25,15){\scriptsize c. Transfer}
                \end{overpic}
            \end{tabular}%
        }
        \end{tcolorbox}
    \end{minipage}
    % \hfill
    \begin{minipage}[c]{0.47\textwidth}
        \caption{
        Controlled same-scan spacing perturbation under increasing through-plane anisotropy.
        A single CT volume is resampled from reference isotropic spacing 
        (1.0$\times$1.0$\times$1.0 mm) to moderate 
        (1.0$\times$1.0$\times$3.0 mm) and higher-anisotropy 
        (1.0$\times$1.0$\times$5.0 mm) variants (top).
        Bottom: VoCo~\cite{wu2026:voco} and Med3DINO are compared using 
        (a) cosine embedding drift,
        (b) feature-space similarity measured by CKA, and 
        (c) cross-spacing semantic transfer, respectively probing embedding movement, preservation of feature-space organization, and semantic transfer under spacing change. %In (c), a classifier trained on reference-spacing features is evaluated on scans at moderate and severe spacings.
        }
        \label{fig:controlled_spacing_visualization}
    \end{minipage}
    \vspace{-1.5em}
\end{figure}

This limitation is especially critical in 3D medical imaging because acquisition geometry directly changes the evidence presented to a volumetric encoder~\cite{joutard2024:hyperspace,LIN2026:anisotropic_medseg_pr}. Voxel spacing, through-plane anisotropy, sampling density, and anatomical coverage can all alter how the same organ appears, even when the underlying anatomy is unchanged. Fig.~\ref{fig:controlled_spacing_visualization} isolates this effect through a controlled same-scan perturbation: a single CT volume is resampled to multiple spacings while the anatomical structure is held fixed. The figure reports three complementary probes: cosine drift measures how far same-scan embeddings move after resampling, CKA~\cite{kornblith2019:CKA} measures whether the feature-space organization is preserved, and transfer measures whether the semantic structure remains usable across spacing variants. 

This setting reveals a gap that downstream evaluation obscures. Compared with Med3DINO-SA, VoCo~\cite{wu2026:voco} exhibits lower cosine drift but weaker preservation of feature-space organization and cross-spacing semantic transfer as anisotropy increases. Low drift alone therefore does not establish robustness: feature organization and label-relevant information must also remain usable. This motivates REFORM-3D to evaluate these properties directly in frozen representations.

To this end, we propose \textbf{REFORM-3D}, a representation-centric evaluation framework for benchmarking frozen representations under controlled variations in acquisition, modality, and anatomy. The framework is organized into three primary modules: \textit{Acquisition Robustness}, which tests whether frozen features remain stable as spacing and anisotropy vary for the same scan; \textit{Cross-Modal Anatomical Alignment}, which tests whether labeled structures remain recoverable across modalities under balanced bidirectional retrieval; and \textit{Anatomical Transfer}, which tests whether organ identity remains recoverable from limited labeled support under case-disjoint evaluation.  Collectively, these modules integrate acquisition sensitivity with semantic and anatomical recoverability into an evidence-based pre-adaptation profile that informs data coverage, geometric pretraining, encoder selection, and targeted downstream validation. Each module uses evidence suited to its representation question, and each claim is scoped to its corresponding \textit{evaluation surface}, defined as the dataset--task--protocol combination.

We further introduce Med3DINO encoders, pretrained on 62,638 (hereafter $62.6$K) unlabeled multimodal 3D volumes under three explicit geometric assumptions: isotropic normalization, voxel-relative sampling, and spacing-aware resampling while holding all other pretraining settings fixed. REFORM-3D reveals lower embedding drift under voxel-relative sampling and stronger anatomical recoverability under spacing-aware resampling, while semantic-transfer advantages depend on the dataset and feature type. Applying the same framework to existing 3D medical encoders reveals distinct strengths across the three modules, showing that representation quality is property-specific rather than reducible to a single model ordering. Our contributions are listed as follows:
\begin{itemize}
    \item We introduce REFORM-3D, a representation-centric framework that evaluates frozen 3D medical encoders through Acquisition Robustness, Cross-Modal Anatomical Alignment, and Anatomical Transfer.
    \item We introduce Med3DINO, a family of 3D encoders pretrained on $62.6$K unlabeled multimodal volumes, including MR, CT, PET, CTA, and CBCT, under three geometric assumptions while other pretraining settings remain fixed.
    \item We benchmark Med3DINO alongside existing 3D medical encoders and find that these models exhibit different strengths in representation stability and semantic or anatomical recoverability.
\end{itemize}

Across heterogeneous datasets and modalities, REFORM-3D evaluates frozen representations under acquisition variation, cross-modal retrieval, and case-disjoint anatomical transfer, providing a reusable basis for studying representation quality beyond downstream scores.

\section{Related Work}
\label{sec:related_work}
\textbf{Self-supervised Medical Vision Foundation Models:} 
% \subsection{Self-supervised Medical Vision Foundation Models}
Self-supervised representation learning in natural images has evolved from contrastive methods such as SimCLR~\cite{chen2020:SimCLR} and MoCo~\cite{he2020:MoCo} to DINO~\cite{caron2021:DINO,oquab2024:dinov2,siméoni:dinov3} pretraining, whose teacher-student formulation has become especially effective for learning dense and local features. This paradigm has since been adapted to medical imaging in both 2D and 3D. In 2D medical imaging, MedDINOv3~\cite{li2025:meddinov3}, Radio-DINO~\cite{zedda2025:radio_dino}, and RAD-DINO~\cite{perez2025:rad_dinov2} demonstrate transferable encoders from DINO-family pretraining on medical images, while DINOv2-based adaptation of ALPNet~\cite{ayzenberg:dinov2_fewshot} supports few-shot medical image segmentation. In 3D imaging, BrainIAC~\cite{tak2026:brainAIC} builds a brain MR foundation model under contrastive self-supervision, while OpenMind~\cite{wald2025:openmind} benchmarks large-scale 3D brain MR self-supervised methods, including SimCLR~\cite{chen2020:SimCLR} and VoCo~\cite{wu2026:voco}. 3DINO~\cite{3dinov2} more directly adapts DINOv2~\cite{oquab2024:dinov2} to volumetric multimodal pretraining, moving the field toward generalist 3D medical encoders. Recent volumetric encoders further span segmentation-oriented transformers and masked-modeling methods with ViT, Mamba, and hybrid architectures, including SAM-Med3D~\cite{samed3d_tnnls}, Hi-End-MAE~\cite{Hi_End_MAE}, MambaMIM~\cite{MambaMIM}, MiM~\cite{MiM}, and GL-MAE~\cite{GL_MAE}. This trajectory also makes clear the need for heterogeneous large-scale pretraining across modalities, anatomical regions, and pathologies if the goal is a general-purpose 3D encoder instead of one confined to a narrow modality or anatomical scope. Nevertheless, these works mainly aim to build stronger encoders through pretraining objectives, architectural adaptation, and scale, while giving much less attention to how frozen features should be evaluated after pretraining. Representation quality is therefore still judged mainly through downstream transfer, not through direct evaluation of what the frozen embedding encodes. This gap is especially relevant for DINO encoders because downstream scores alone do not fully characterize their dense and local features, particularly in 3D imaging, where acquisition geometry changes how the same anatomy is presented. Existing benchmarks address medical vision-language performance~\cite{zhou2024:benchx,chen2024:gmai_bench} and fairness across foundation models~\cite{jin2024:fairmedfm}. Our focus is acquisition sensitivity and anatomical recoverability in frozen 3D representations. 

\textbf{Medical Foundation Model Benchmarks:} 
% \subsection{Medical Foundation Model Benchmarks}
There have been several efforts to benchmark medical foundation models across tasks, datasets, fairness criteria, and deployment-oriented settings~\cite{zhou2024:benchx,jin2024:fairmedfm,chen2024:gmai_bench,gai2025:3d_rad}. Many of these efforts focus on medical Vision-Language Models (VLMs)~\cite{zhou2024:benchx,chen2024:gmai_bench,gai2025:3d_rad}, while FairMedFM also evaluates vision-only models~\cite{jin2024:fairmedfm}. BrainIAC~\cite{tak2026:brainAIC} and OpenMind~\cite{wald2025:openmind} emphasize predominantly brain-MRI pretraining. These works have made model comparison more systematic through shared task definitions, evaluation protocols, and reporting conventions. REFORM-3D complements these benchmarks by jointly evaluating acquisition sensitivity, cross-modal anatomical alignment, and case-disjoint anatomical transfer in frozen 3D encoders. In 3D medical imaging, that distinction matters because acquisition geometry changes how the same anatomy is presented to the encoder.

\textbf{Downstream Robustness vs. Representation Evaluation:}
% \subsection{Downstream Robustness vs. Representation Evaluation}
In 3D medical imaging, anisotropy-aware and cross-slice modeling show that acquisition geometry affects the evidence presented to a volumetric model~\cite{joutard2024:hyperspace,LIN2026:anisotropic_medseg_pr,cobo2025physical:anisotropy,hung2024:csam_cross_slice}. When spacing or slice thickness changes, through-plane resolution and cross-slice continuity also change, so the same organ is no longer presented under the same evidence. Existing studies usually examine acquisition variation after task-specific adaptation. That indicates anisotropy matters, but it does not determine whether sensitivity originates in the frozen representation itself or in the adapted task pipeline built on top of it. Our framework evaluates this distinction by examining frozen 3D features through stability under acquisition variation, retrieval of labeled structures across modalities, and case-disjoint anatomical transfer.

% \vspace{-0.8em}
\section{The REFORM-3D Framework}
\label{sec:benchmark_design}
\begin{figure*}[!t]
    \centering
    \includegraphics[width=\textwidth]{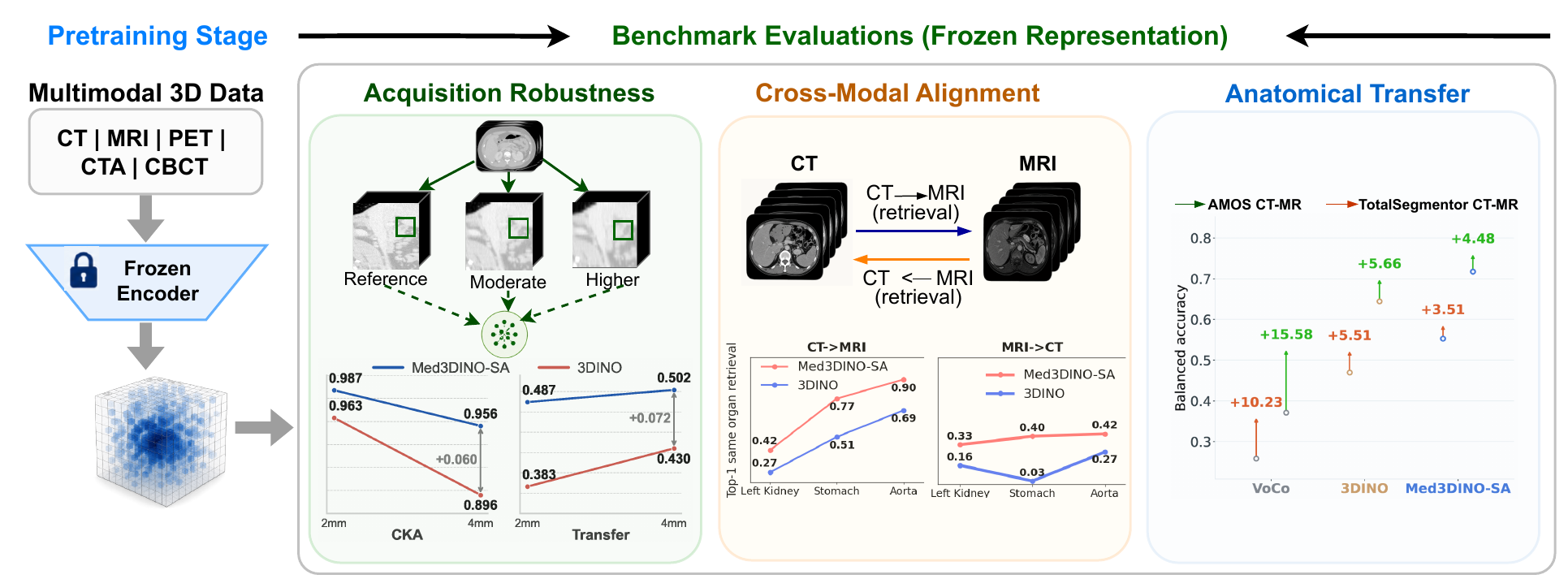}
    \caption{\textbf{Overview of REFORM-3D Framework}. The framework assesses frozen 3D representations along three axes: (a) Acquisition Robustness, using controlled same-scan spacing perturbations to measure stability, structural preservation (CKA), and transfer; (b) Cross-Modal Anatomical Alignment, evaluating bidirectional cross-modal retrieval; and (c) Anatomical Transfer from limited labels under case-disjoint evaluation. Plots illustrate main trends across models and metrics.}
    \label{fig:main_framework}
    \vspace{-1.5em}
\end{figure*}

The goal of REFORM-3D is to evaluate what frozen 3D representations encode, rather than how well they perform after adaptation. In 3D medical imaging, acquisition variability, modality differences, and anatomical diversity can introduce shortcut features that are not anatomically grounded. Hence, REFORM-3D evaluates frozen representations through three complementary modules: Acquisition Robustness under controlled same-scan spacing changes, Cross-Modal Anatomical Alignment through balanced bidirectional retrieval, and Anatomical Transfer using limited support and case-disjoint queries. These protocols separately characterize acquisition sensitivity, cross-modal structure recovery, and anatomical recovery. Fig.~\ref{fig:main_framework} summarizes the framework.

\subsection{Pretraining Corpus and Geometric Regimes.} We use REFORM-3D to evaluate Med3DINO, a family of volumetric DINOv2 encoders~\cite{oquab2024:dinov2,3dinov2} pretrained on $62.6$K unlabeled CT, MR, PET, CTA, and CBCT volumes covering over 180 anatomical structures. The family comprises three geometric pretraining regimes: isotropic normalization (ISO), voxel-relative sampling (REL), and spacing-aware resampling (SA) while fixing the corpus, initialization, architecture, objective, non-geometric augmentations, and optimization budget. This controlled design isolates the complete geometric policy within Med3DINO; other pretrained encoders serve as contextual representation baselines. Further corpus and preprocessing details are provided in Supplementary Section~\ref{sec:pretraining_details}. % and~\ref{subsec:pretraining_preprocessing}.
% \vspace{-1.0em}

\subsection{REFORM-3D Modules}
\label{subsec:benchmark_modules}
REFORM-3D is organized into three modules, each defined by a distinct representation-level question about frozen 3D features. The encoder remains frozen throughout, and each module uses metrics matched to its evaluation setting.

\textbf{Acquisition Robustness.}
This module asks whether changing voxel spacing alters a frozen representation when the same scan is held fixed. Let $x_i$ denote a scan, let $f_\theta$ be a frozen encoder, and let $T(x_i; s)$ denote deterministic preprocessing followed by controlled resampling of $x_i$ to target spacing $s$. The controlled evaluation therefore compares embeddings \(z_i(s)=f_\theta(T(x_i; s))\) across multiple spacing variants of the same scan, so that spacing is the intended source of variation. We evaluate this controlled setting in three complementary ways. The first is within-scan cosine drift : \(d_i(a,b) = 1 - \cos\left(z_i^{(a)}, z_i^{(b)}\right)\),
% \begin{equation}
% d_i(a,b) = 1 - \cos\left(z_i^{(a)}, z_i^{(b)}\right),
% \end{equation}
which measures how far the representation of a scan moves when spacing alone is perturbed. The second is cross-variant CKA~\cite{kornblith2019:CKA}, computed over the same scans to assess preservation of feature-space organization. The third is semantic transfer across spacing variants, $\mathrm{SemTrans}(a \rightarrow b)$, which asks whether a classifier trained on features at spacing $a$ can recover the corresponding dataset-specific semantic targets from features at spacing $b$. We fix cases, targets, five-fold splits, and classifier settings, fitting the standardizer and classifier only at the source spacing and applying them unchanged at the target spacing. Hence, these metrics collectively distinguish embedding movement, feature-space reorganization, and reduced semantic recoverability under controlled resampling. % rather than scanner physics.

\textbf{Cross-Modal Anatomical Alignment.}  
This module evaluates whether labeled structures remain recoverable across modalities under balanced bidirectional retrieval. Its primary evaluation uses organ-level CT--MR retrieval. For a query $q$, let $C(q)$ be an organ-balanced candidate pool from the opposite modality, and let $P(q) \subseteq C(q)$ denote same-organ positives. Retrieval is evaluated in both CT$\rightarrow$MR and MR$\rightarrow$CT directions. Candidates are ranked by cosine similarity. We compute average precision (AP) per query, average it within each direction, and then equally across directions:
\begin{equation}
    \begin{aligned}
        \mathrm{AP}(q) &=\frac{1}{|P(q)|}\sum_{k=1}^{|C(q)|}\mathrm{Prec}@k(q)\,\mathbf{1}\!\left[c_k\in P(q)\right], \qquad \mathrm{mAP}_{d}=\frac{1}{|Q_d|}\sum_{q\in Q_d}\mathrm{AP}(q),\\
        \mathrm{mAP}_{\leftrightarrow} &=\frac{1}{2}\sum_d \mathrm{mAP}_{d}, \qquad d\in\{\mathrm{CT}\!\rightarrow\!\mathrm{MR}, \mathrm{MR}\!\rightarrow\!\mathrm{CT}\}.
    \end{aligned}
\end{equation}
where \(c_k\) is the candidate at rank \(k\), \(Q_d\) is the query set for direction \(d\), and \(\mathrm{mAP}_{\leftrightarrow}\) is the equal-weight bidirectional mean. We additionally report Top-1 and Top-5 accuracy. Hence, these metrics measure whether same-organ cross-modal matches are ranked above competing organs, without implying voxel-level alignment or full modality invariance.

\textbf{Anatomical Transfer.}
This module evaluates whether organ identity can be recovered from frozen features using limited labeled support. The primary setting uses two support samples per organ and modality, while query samples are drawn from cases disjoint from all support cases; the encoder remains frozen. We use balanced accuracy (BA) as the common recoverability metric,
\begin{equation}
    \begin{aligned}
        \mathrm{BA}=\frac{1}{|\mathcal{O}|}\sum_{o\in\mathcal{O}}\mathrm{Recall}(o)
    \end{aligned}
    \label{eqn.BA}
\end{equation}
% $\mathrm{BA}=\frac{1}{|\mathcal{O}|}\sum_{o\in\mathcal{O}}\mathrm{Recall}(o)$, 
where $\mathcal{O}$ is the set of evaluated organs and $\mathrm{Recall}(o)$ is recall on the case-disjoint queries for organ $o$. Feature standardization is fitted on the support set and applied unchanged to the query set. As the nearest-centroid frozen-feature baseline, we form one support centroid per organ class,
\begin{equation}
    \begin{aligned}
        c_o=\frac{1}{|S_o|}\sum_{z\in S_o}z,
    \end{aligned}
\end{equation}
where $S_o$ contains the standardized support embeddings for organ $o$ across modalities. Each standardized query embedding is assigned to the organ class whose centroid $c_o$ is nearest in Euclidean distance, yielding $\mathrm{BA}_{\mathrm{centroid}}$. We then train a lightweight linear probe on the same balanced support set and report $\mathrm{BA}_{\mathrm{probe}}$, together with
$\Delta_{\mathrm{adapt}}=\mathrm{BA}_{\mathrm{probe}}-\mathrm{BA}_{\mathrm{centroid}}$. These quantities distinguish frozen organ separability, the gain from limited adaptation, and organ recoverability; they do not imply that the evaluated organs were absent during pretraining.
\begin{table*}[!t]
    \centering
    \small
    \caption{Evaluation datasets grouped by inclusion in the Med3DINO pretraining corpus. Exposure is defined at the dataset level and does not imply that individual anatomical classes were unseen.}
    \label{tab:benchmark_dataset_roles}
    \resizebox{\textwidth}{!}{%
    \begin{tabular}{p{0.20\textwidth} p{0.44\textwidth} p{0.3\textwidth}}
        \toprule
        Pretraining exposure & Dataset & Evaluation role \\
        \midrule
        \rowcolor{blue!8}
        Seen in pretraining &
        AbdomenAtlas~\cite{li2024:SuPreM};
        ImageCAS~\cite{zeng2023:imagecas};
        JHU-Stroke Brain MR~\cite{liu2023:jhu_stroke};
        TotalSegmentator-CT~\cite{wasserthal2023:totalsegmentatorCT};
        HECKTOR-2025~\cite{hecktor2025}; &
        Acquisition Robustness; CT contribution to CT--MR alignment and anatomical transfer; CT--PET lesion-category retrieval\\
        % Spacing sensitivity; in-corpus anatomical consistency; CT–PET lesion-category retrieval \\
        \midrule
        \rowcolor{green!8}
        Unseen in pretraining &
        KiTS23~\cite{heller2023kits21};
        TotalSegmentator-MR~\cite{akinci:totalsegmentatorMR};
        CHAOS~\cite{kavur2021:chaos};
        AMOS22~\cite{ji2022:AMOS2022};
        MMWHS~\cite{zhuang2019:mmwhs_ct_mr} &
        Acquisition Robustness; CT--MR alignment; anatomical transfer \\
        \bottomrule
    \end{tabular}%
    }
    \vspace{-1.5em}
\end{table*}

\subsection{Evaluation Datasets and Surfaces}
\label{secsub:benchmark_evaluation_dataset}
REFORM-3D does not use one pooled benchmark cohort. Instead, each module uses the evaluation surface required by its scientific question: controlled same-scan spacing perturbation for Acquisition Robustness, balanced cross-modal retrieval for Cross-Modal Anatomical Alignment, and case-disjoint organ classification for Anatomical Transfer. Table~\ref{tab:benchmark_dataset_roles} groups datasets by their inclusion in the Med3DINO pretraining corpus. These labels describe dataset-level exposure for Med3DINO, not anatomical classes absent from pretraining, and do not automatically apply to external encoders. %Detailed construction and filtering rules are provided in Supplementary Section~\ref{sec:benchmark_dataset_construction}.

\textbf{Acquisition Robustness} uses AbdomenAtlas~\cite{li2024:SuPreM}, KiTS23~\cite{heller2023kits21}, TotalSegmentator-MR~\cite{akinci:totalsegmentatorMR}, ImageCAS~\cite{zeng2023:imagecas}, and JHU-Stroke Brain MR~\cite{liu2023:jhu_stroke}, each converted into controlled spacing variants of the same scan. This module therefore spans seen surfaces (AbdomenAtlas, ImageCAS, and JHU-Stroke Brain MR) as well as pretraining-unseen surfaces (KiTS23 and TotalSegmentator-MR) while preserving the same evidential core: spacing changes while scan identity and anatomy are fixed. Controlled same-scan comparisons carry the main robustness claim; native-spacing analyses provide observational support because they compare different scans with different spacing ranges.

\textbf{Cross-Modal Anatomical Alignment} uses TotalSegmentator CT--MR~\cite{wasserthal2023:totalsegmentatorCT,akinci:totalsegmentatorMR}, MMWHS~\cite{zhuang2019:mmwhs_ct_mr}, and AMOS22~\cite{ji2022:AMOS2022} as balanced multi-organ CT--MR surfaces; CHAOS~\cite{kavur2021:chaos} is reported separately as a liver-only surface. TotalSegmentator CT--MR has mixed Med3DINO exposure, whereas MMWHS, AMOS22, and CHAOS are pretraining-unseen datasets. We additionally evaluate HECKTOR-2025~\cite{hecktor2025} CT--PET using primary-tumor and metastatic-lymph-node retrieval with same-patient targets excluded. Because this surface uses lesion categories, it is interpreted separately from organ-level CT--MR retrieval.

\textbf{Anatomical Transfer} uses TotalSegmentator CT--MR, MMWHS, and AMOS22 with two support samples per organ and modality and case-disjoint queries. TotalSegmentator CT--MR is a mixed-exposure surface, while MMWHS and AMOS22 are pretraining-unseen datasets. These surfaces evaluate organ recoverability from limited labeled support and do not imply that the evaluated organ classes were absent during pretraining.
\begin{table}[t]
    \centering
    \scriptsize
    \setlength{\tabcolsep}{3pt}
    \caption{Evaluated frozen encoders and their pretraining settings. External encoders provide contextual comparisons; the Med3DINO triplet controls other pretraining settings to compare geometric policies. GAP and CLS denote global average pooling over token or spatial positions and class token, respectively; $^\dagger$Predominantly MR or CT/MR with a small PET subset.}
    \label{tab:fm_comparison}
    \resizebox{\textwidth}{!}{%
    \begin{tabular}{@{}lllcrll@{}}
        \toprule
        Model & Encoder & Evaluated feature(s) & Modalities & Volumes & Pretraining objective & Pretraining geometry \\
        \midrule
        \multicolumn{7}{l}{\cellcolor{softgreen}\textit{External Encoders}} \\
        BrainIAC~\cite{tak2026:brainAIC} & ViT-B & First patch token & MR & 32.0K & SimCLR & Isotropic normalization \\
        OpenMind~\cite{wald2025:openmind} & Primus-M & GAP & MR/PET$^\dagger$ & 113.9K & Volume contrastive & Isotropic normalization \\
        VoCo~\cite{wu2026:voco} & SwinUNETR & Multiscale GAP & CT & $\sim$160K & Volume contrastive & Region-specific resampling \\
        3DINO~\cite{3dinov2} & ViT-L & CLS / GAP & CT/MR/PET$^\dagger$ & 98.8K & DINOv2 & Voxel-relative sampling \\
        SAM-Med3D~\cite{samed3d_tnnls} & ViT-B & Post-neck GAP & CT/MR & $\sim$22K& Supervised Dice + CE & Isotropic normalization \\
        Hi-End-MAE~\cite{Hi_End_MAE} & ViT & CLS & CT & 10.0K & Hierarchical MAE & Isotropic normalization \\
        MambaMIM~\cite{MambaMIM} & CNN--Mamba & Final-stage GAP & CT & 6.8K & Masked reconstruction & Isotropic normalization \\
        MiM~\cite{MiM} & ConvViT-B & GAP & CT & 10.5K & MAE + InfoNCE & Multiscale voxel sampling \\
        GL-MAE~\cite{GL_MAE} & ViT-B & CLS & CT & 0.75K & MAE + consistency & Multiscale voxel sampling \\
        \midrule
        \multicolumn{7}{l}{\cellcolor{softgreen}\textit{Controlled Med3DINO Family}} \\
        Med3DINO-ISO & ViT-L & CLS / GAP & CT/MR/PET/CTA/CBCT & 62.6K & DINOv2 & Isotropic normalization \\
        Med3DINO-REL & ViT-L & CLS / GAP & CT/MR/PET/CTA/CBCT & 62.6K & DINOv2 & Voxel-relative sampling \\
        Med3DINO-SA & ViT-L & CLS / GAP & CT/MR/PET/CTA/CBCT & 62.6K & DINOv2 & Spacing-aware resampling \\
        \bottomrule
    \end{tabular}%
    }
    \vspace{-1.2em}
\end{table}

% \vspace{-1.4em}
\section{Benchmark Results}
\label{sec:benchmark_results}
We evaluate frozen 3D representations with the three REFORM-3D benchmark modules: Acquisition Robustness, Cross-Modal Anatomical Alignment, and Anatomical Transfer. We first examine sensitivity to acquisition geometry, followed by cross-modal retrieval and anatomical recoverability across modalities. Each conclusion is restricted to the evidence supported by its evaluation setting. Evaluation dataset construction, metric definitions, and implementation details are provided in Supplementary Sections~\ref{sec:benchmark_dataset_construction}, \ref{sec:benchmark_protocol_metric}, and~\ref{sec:implementation_reproducibility_benchmarking}.

\textbf{Evaluated Models.}
We evaluate Med3DINO alongside of nine frozen encoders, including BrainIAC, OpenMind, VoCo, 3DINO, SAM-Med3D, Hi-End-MAE, MambaMIM, MiM, and GL-MAE using their designated feature interfaces. Within the controlled Med3DINO comparison, ISO is pretrained after resampling volumes to a fixed isotropic grid. REL is pretrained on anisotropic volumes, but defines crop scale relative to the current voxel grid. SA is pretrained on anisotropic volumes while defining the through-plane crop extent in physical units and using anisotropy-aware rotation handling. With the remaining pretraining settings fixed, this triplet isolates the complete geometric policy within Med3DINO. External encoders provide contextual comparisons across different pretraining designs. Table~\ref{tab:fm_comparison} summarizes the evaluated models.

\subsection{Acquisition Robustness Results}
\label{subsec:acquisition_robutst_results}
Acquisition Robustness evaluates whether spacing changes the frozen representation when scan identity is held fixed. Table~\ref{tab:bench_acquisition_controlled} is the primary evidence: drift measures feature displacement, CKA measures preservation of feature-space structure, and semantic transfer measures whether label-relevant structure remains usable across spacing variants. Together, these metrics show that the controlled same-scan results do not support one global robustness ranking. Within the controlled ISO/REL/SA comparison, the reported CLS results give REL the lowest drift and SA the highest semantic transfer on three of the four displayed datasets. ImageCAS instead favors ISO for drift and REL for transfer. CKA varies by dataset and feature type. Among the external encoders, MiM and GL-MAE retain high CKA, but SAM-Med3D achieves higher mean transfer than either on KiTS23 and TS-MR. We further provide JHU-Stroke results and paired ISO/REL/SA comparisons in supplementary Tables~\ref{tab:bench_acquisition_controlled_seeded_full} and~\ref{tab:acquisition_paired_analysis}, respectively. Classifier, semantic-target, and preprocessing sensitivities are reported in Supplementary Sections~\ref{subsec:classifier_sensitivity}, \ref{subsec:semantic_target_sensitivity}, and~\ref{subsec:protocol_stability}.

\begin{figure*}[!t]
    \centering
    \includegraphics[width=\textwidth]{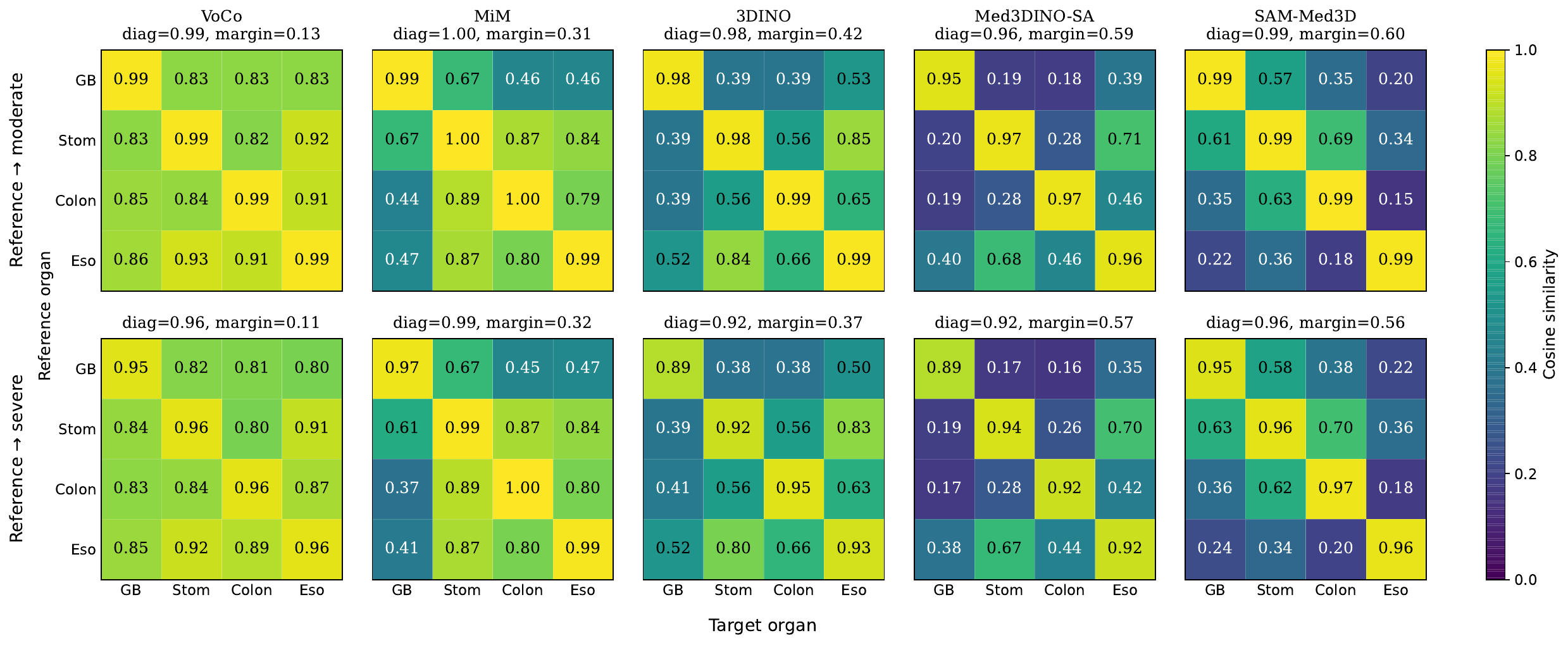}
    \caption{Organ-specific feature separation under controlled same-scan spacing perturbations in AbdomenAtlas. Heatmaps show cosine similarity between dense features pooled within organ masks at reference spacing ($1\times1\times1$ mm) and moderate ($1\times1\times3$ mm; top) or severe ($1\times1\times5$ mm; bottom) target spacings. Diagonal entries compare the same organ; off-diagonal entries compare different organs. The margin is mean diagonal minus mean off-diagonal similarity, distinguishing organ-specific correspondence from uniformly high similarity.}
    \label{fig:organ_attention_map}
    \vspace{-1.5em}
\end{figure*}

Fig.~\ref{fig:organ_attention_map} further provides a local feature-space view of acquisition robustness. VoCo~\cite{wu2026:voco} retains uniformly high cosine similarity but weak diagonal--off-diagonal separation. MiM~\cite{MiM} also maintains high same-organ similarity, while 3DINO~\cite{3dinov2} shows greater organ-specific separation. SAM-Med3D~\cite{samed3d_tnnls} and Med3DINO-SA exhibit larger displayed margins under both perturbation conditions. These illustrative profiles distinguish high same-organ similarity from separation between anatomical structures.
\begin{table*}[!t]
    \centering
    \scriptsize
    \setlength{\tabcolsep}{2.5pt}
    \caption{Controlled same-scan acquisition robustness under spacing perturbations. Reported results are mean$\pm$SD over three-seed evaluations. External encoders provide contextual comparisons; ISO/REL/SA form the controlled geometric-policy comparison. 3DINO and Med3DINO use CLS and GAP; other feature interfaces follow Table~\ref{tab:fm_comparison}. TS-MR denotes TotalSegmentator-MR.}

    \label{tab:bench_acquisition_controlled}
    \resizebox{\textwidth}{!}{%
    \begin{tabular}{@{}l@{\hspace{6pt}}rrr@{\hspace{8pt}}rrr@{\hspace{8pt}}rrr@{\hspace{8pt}}rrr@{}}
        \toprule
        Method & \multicolumn{3}{c}{AbdomenAtlas~\cite{li2024:SuPreM}} & \multicolumn{3}{c}{KiTS23~\cite{heller2023kits21}} & \multicolumn{3}{c}{ImageCAS~\cite{zeng2023:imagecas}} & \multicolumn{3}{c}{TS-MR~\cite{akinci:totalsegmentatorMR}} \\
        \cmidrule(lr){2-4} \cmidrule(lr){5-7} \cmidrule(lr){8-10} \cmidrule(lr){11-13}
        & Drift$\downarrow$ & CKA$\uparrow$ & Trans$\uparrow$ & Drift$\downarrow$ & CKA$\uparrow$ & Trans$\uparrow$ & Drift$\downarrow$ & CKA$\uparrow$ & Trans$\uparrow$ & Drift$\downarrow$ & CKA$\uparrow$ & Trans$\uparrow$ \\
        \midrule
        \multicolumn{13}{l}{\cellcolor{lightgraybg}\textbf{External encoders}} \\
        BrainIAC~\cite{tak2026:brainAIC}  & \ul{0.30}$\pm0.02$ & 97.60$\pm0.17$ & 67.20$\pm1.92$ & \textbf{0.30}$\pm0.00$ & 96.70$\pm0.00$ & 46.60$\pm4.89$ & 0.90$\pm0.00$ & 93.90$\pm0.00$ & 52.50$\pm0.59$ & 0.50$\pm0.00$ & 99.10$\pm0.00$ & 55.70$\pm2.43$ \\
        OpenMind~\cite{wald2025:openmind} & \textbf{0.10}$\pm0.05$ & 77.70$\pm2.96$ & 57.10$\pm2.05$ & 1.20$\pm0.26$ & 80.70$\pm7.12$ & 46.70$\pm2.98$ & \textbf{0.20}$\pm0.18$ & 53.00$\pm4.98$ & 50.10$\pm0.16$ & 0.20$\pm0.01$ & 92.60$\pm3.65$ & 49.30$\pm1.84$ \\
        VoCo~\cite{wu2026:voco}           & 0.80$\pm0.09$ & 86.70$\pm0.10$ & 74.60$\pm0.15$ & 0.70$\pm0.00$ & 81.30$\pm0.00$ & 46.10$\pm4.07$ & 2.30$\pm0.00$ & 42.60$\pm0.00$ & 52.00$\pm0.13$ & 0.10$\pm0.00$ & 95.94$\pm0.00$ & 71.10$\pm0.82$ \\
        3DINO (CLS)~\cite{3dinov2}        & 16.30$\pm0.01$ & 87.50$\pm0.07$ & 75.10$\pm0.18$ & 15.40$\pm0.00$ & 77.30$\pm0.00$ & 45.50$\pm1.08$ & 32.10$\pm0.00$ & 83.40$\pm0.00$ & \textbf{55.90}$\pm0.01$ & 6.00$\pm0.00$ & 92.40$\pm0.00$ & 74.60$\pm0.73$ \\
        3DINO (GAP)~\cite{3dinov2}        & 3.60$\pm0.00$ & 96.20$\pm0.11$ & 77.80$\pm0.59$ & 3.70$\pm0.00$ & 91.90$\pm0.00$ & 39.90$\pm2.23$ & 11.30$\pm0.00$ & 88.30$\pm0.00$ & 53.90$\pm0.35$ & 1.30$\pm0.00$ & 98.90$\pm0.00$ & 80.70$\pm0.23$ \\
        SAM-Med3D~\cite{samed3d_tnnls}    & 3.35$\pm0.04$ & 86.88$\pm0.37$ & 74.66$\pm0.32$ & 2.83$\pm0.00$ & 95.27$\pm0.00$ & \ul{52.82}$\pm5.00$ & \ul{0.87}$\pm0.00$ & 96.13$\pm0.00$ & 54.23$\pm0.49$ & 0.51$\pm0.00$ & 97.53$\pm0.00$ & 72.61$\pm1.40$ \\
        Hi-End-MAE~\cite{Hi_End_MAE}      & 0.94$\pm0.01$ & 97.18$\pm0.17$ & 72.68$\pm0.67$ & 0.78$\pm0.00$ & 97.44$\pm0.00$ & 47.99$\pm3.40$ & 1.15$\pm0.00$ & 93.61$\pm0.00$ & 51.87$\pm0.14$ & 0.23$\pm0.00$ & 99.18$\pm0.00$ & 70.28$\pm1.85$ \\
        MambaMIM~\cite{MambaMIM}          & 5.49$\pm0.05$ & 90.44$\pm0.38$ & 72.17$\pm0.54$ & 5.58$\pm0.00$ & 94.62$\pm0.00$ & 50.26$\pm2.33$ & 14.38$\pm0.00$ & 83.53$\pm0.00$ & 51.38$\pm0.11$ & 0.13$\pm0.00$ & 98.97$\pm0.00$ & 71.25$\pm0.43$ \\
        MiM~\cite{MiM}                    & 1.44$\pm0.01$ & \textbf{99.59}$\pm0.02$ & 73.69$\pm0.49$ & 0.75$\pm0.00$ & \ul{ 99.86}$\pm0.00$ & 48.94$\pm2.62$ & 2.06$\pm0.00$ & \ul{99.20}$\pm0.00$ & 54.01$\pm0.38$ & \textbf{0.01}$\pm0.00$ & \ul{99.94}$\pm0.00$ & 56.35$\pm1.53$ \\
        GL-MAE~\cite{GL_MAE}              & 0.78$\pm0.00$ & \ul{99.29}$\pm0.01$ & 74.65$\pm0.94$ & \ul{0.50}$\pm0.00$ & \textbf{99.88}$\pm0.00$ & 48.51$\pm4.69$ & 1.06$\pm0.00$ & \textbf{99.35}$\pm0.00$ & 52.79$\pm1.02$ & \ul{0.04}$\pm0.00$ & \textbf{99.95}$\pm0.00$ & 58.70$\pm0.32$ \\
        \midrule
        \multicolumn{13}{l}{\cellcolor{lightgraybg}\textbf{Controlled Med3DINO: CLS}} \\
        Med3DINO-ISO        & 19.00$\pm0.09$ & 78.50$\pm0.14$ & 73.60$\pm0.78$ & 20.00$\pm0.00$ & 73.60$\pm0.00$ & 34.80$\pm4.02$ & 22.00$\pm0.00$ & 93.10$\pm0.00$ & 52.90$\pm0.57$ & 7.00$\pm0.00$ & 92.40$\pm0.00$ & 59.30$\pm0.34$ \\
        Med3DINO-REL        & 10.90$\pm0.07$ & 89.10$\pm0.13$ & 77.20$\pm0.61$ & 12.00$\pm0.00$ & 86.10$\pm0.00$ & 44.10$\pm2.22$ & 24.20$\pm0.00$ & 83.30$\pm0.00$ & \ul{54.60}$\pm0.45$ & 5.60$\pm0.00$ & 93.50$\pm0.00$ & 77.00$\pm0.91$ \\
        Med3DINO-SA         & 18.40$\pm0.06$ & 85.30$\pm0.05$ & \textbf{81.00}$\pm0.16$ & 19.20$\pm0.00$ & 84.30$\pm0.00$ & \textbf{57.80}$\pm3.25$ & 35.60$\pm0.00$ & 83.80$\pm0.00$ & 54.20$\pm0.37$ & 5.70$\pm0.00$ & 94.20$\pm0.00$ & 80.60$\pm1.18$ \\
        \midrule
        \multicolumn{13}{l}{\cellcolor{lightgraybg}\textbf{Controlled Med3DINO: GAP}} \\
        Med3DINO-ISO        & 4.10$\pm0.02$ & 95.90$\pm0.11$ & 72.50$\pm0.15$ & 3.90$\pm0.00$ & 91.50$\pm0.00$ & 43.80$\pm2.41$ & 13.20$\pm0.00$   & 92.40$\pm0.00$ & 54.00$\pm0.23$ & 2.00$\pm0.00$ & 97.90$\pm0.00$ & 78.00$\pm0.87$ \\
        Med3DINO-REL        & 1.90$\pm0.03$ & 96.70$\pm0.06$ & 78.90$\pm0.78$ & 4.40$\pm0.00$ & 92.60$\pm0.00$ & 41.60$\pm3.55$ & 12.30$\pm0.00$   & 80.90$\pm0.00$ & 53.40$\pm0.43$ & 1.80$\pm0.00$ & 97.40$\pm0.00$ & \ul{81.70}$\pm1.05$ \\
        Med3DINO-SA         & 5.40$\pm0.02$ & 97.00$\pm0.02$ & \ul{79.70}$\pm0.43$ & 5.80$\pm0.00$ & 94.30$\pm0.00$ & 49.20$\pm3.32$ & 9.70$\pm0.00$   & 85.80$\pm0.00$ & 53.70$\pm0.24$ & 1.70$\pm0.00$ & 99.20$\pm0.00$ & \textbf{83.50}$\pm0.57$ \\
        \bottomrule
    \end{tabular}%
    }
    % \vspace{-2.4em}
\end{table*}

Native-spacing results in Table~\ref{tab:bench_acquisition_native} provide supporting, but not primary, evidence because they compare different scans rather than controlled variants of the same scan. Even in this observational setting, Med3DINO-SA gives the highest semantic transfer for both CLS-token and patch-token features across the three datasets, while CKA remains more feature- and dataset-dependent. These analyses distinguish embedding stability from semantic recoverability, rather than establishing one acquisition-robustness ranking across datasets and features.
\begin{table}[!t]
	\centering
	\scriptsize
    \caption{Native-spacing observational analysis across spacing groups. CKA measures cross-group feature-space similarity, and semantic transfer (Trans) measures whether label-relevant structure remains usable across native-spacing differences. The first group uses CLS for 3DINO and Med3DINO; other interfaces follow Table~\ref{tab:fm_comparison}. }
    % \resizebox{\textwidth}{!}{
    \begin{tabular}{lrrrrrr}
        \toprule
        Method & \multicolumn{2}{c}{AbdomenAtlas~\cite{li2024:SuPreM}} & \multicolumn{2}{c}{KiTS23~\cite{heller2023kits21}} & \multicolumn{2}{c}{TotalSegmentator-MR~\cite{akinci:totalsegmentatorMR}} \\
        \cmidrule(lr){2-3} \cmidrule(lr){4-5} \cmidrule(lr){6-7} & CKA$\uparrow$  & Trans$\uparrow$ & CKA$\uparrow$ & Trans$\uparrow$ & CKA$\uparrow$  & Trans$\uparrow$ \\
        \midrule
        % \multicolumn{7}{l}{\cellcolor{lightgraybg}\textbf{CLS-token Features}} \\
        \multicolumn{7}{l}{\cellcolor{lightgraybg}\textbf{Model-Specific Features (CLS for 3DINO and Med3DINO)}} \\
        \addlinespace[2pt]
        BrainIAC~\cite{tak2026:brainAIC}      & 0.50   & 61.20       & 2.70            & 45.90            & 2.70       & 51.60 \\
        OpenMind~\cite{wald2025:openmind}     & 0.30    & 52.30      & 3.70            & 51.00            & 4.00       & 48.60 \\
        VoCo~\cite{wu2026:voco}               & 0.60    & 66.10      & 6.10            & 49.90            & 6.60       & 59.70 \\
        3DINO~\cite{3dinov2}                  & 14.10           & 66.40            & \ul{52.00}       & 49.80            & \textbf{49.20}   & \ul{65.90} \\
        Med3DINO-ISO                          & \textbf{16.20}  & 65.90            & \textbf{54.70}   & 47.70            & \ul{45.00}       & 56.40 \\
        Med3DINO-REL                          & 13.70           & \ul{67.10}       & 48.40            & \ul{54.20}       & 38.90            & 65.60 \\
        Med3DINO-SA                           & \ul{15.60}      & \textbf{71.00}   & 50.80            & \textbf{56.40}   & 35.20            & \textbf{67.20} \\
        \midrule
        \multicolumn{7}{l}{\cellcolor{lightgraybg}\textbf{Average-Pooled Patch-token Features}} \\
        \addlinespace[2pt]
        3DINO~\cite{3dinov2}                  & \ul{13.20}       & 68.30            & 11.80            & 52.20            & \ul{12.80}       & 66.20 \\
        Med3DINO-ISO                          & 5.30            & \ul{69.50}       & 13.90            & \ul{54.70}       & 10.90            & 60.90 \\
        Med3DINO-REL                          & 12.70            & 67.10            & \ul{17.20}       & 54.50            & \textbf{13.90}   & \ul{66.60} \\
        Med3DINO-SA                           & \textbf{13.30}   & \textbf{72.30}   & \textbf{18.40}   & \textbf{59.60}   & 12.70            & \textbf{69.30} \\
        \bottomrule
    \end{tabular}
    % }
    \label{tab:bench_acquisition_native}
    % \vspace{-2.8em}
\end{table}

\subsection{Cross-Modal Anatomical Alignment Results}
\label{subsec:cross_modal_anatomical_alignment_results}
This module evaluates same-organ CT--MR recoverability under balanced bidirectional retrieval in frozen feature space. Table~\ref{tab:bench_alignment_results} reports mAP, Top-1, and Top-5, averaging the two retrieval directions. Within the controlled Med3DINO comparison, SA achieves the highest scores across all three metrics on TotalSegmentator CT--MR, MMWHS, and AMOS22 with both CLS and GAP features. GAP further increases SA's retrieval scores relative to CLS on these datasets. Among the added encoders, SAM-Med3D has the highest mAP on TotalSegmentator and AMOS22, whereas Hi-End-MAE leads on MMWHS, revealing dataset-dependent retrieval strengths. CHAOS results are interpreted separately because it evaluates identifier-based pairing between different CT and MR patients for liver retrieval rather than multi-organ discrimination. These results support stronger cross-modal organ recoverability under spacing-aware pretraining within the controlled recipe, using crops that retain surrounding context. % Complete results are provided in Supplementary Table~\ref{tab:bench_alignment_seeded_full}.
\begin{table*}[!t]
    \centering
    \scriptsize
    \setlength{\tabcolsep}{3pt}
    \caption{Frozen bidirectional CT--MR retrieval. Results are reported as mean$\pm$SD over three-seed evaluations. 3DINO and Med3DINO use CLS and GAP; other feature interfaces follow Table~\ref{tab:fm_comparison}.} % CHAOS uses patient-ID-matched liver retrieval, not multi-organ retrieval.}
    \label{tab:bench_alignment_results}
    \resizebox{\textwidth}{!}{%
    \begin{tabular}{@{}lrrr@{\hspace{8pt}}rrr@{\hspace{8pt}}rrr@{\hspace{8pt}}rrr@{}}
        \toprule
        Method & \multicolumn{3}{c}{TotalSeg CT--MR~\cite{akinci:totalsegmentatorMR,wasserthal2023:totalsegmentatorCT}} & \multicolumn{3}{c}{MMWHS~\cite{zhuang2019:mmwhs_ct_mr}} & \multicolumn{3}{c}{AMOS22~\cite{ji2022:AMOS2022}} & \multicolumn{3}{c}{CHAOS~\cite{kavur2021:chaos} (liver-only)} \\
        \cmidrule(lr){2-4} \cmidrule(lr){5-7} \cmidrule(lr){8-10} \cmidrule(lr){11-13}
         & mAP & Top-1 & Top-5 & mAP & Top-1 & Top-5 & mAP & Top-1 & Top-5 & mAP & Top-1 & Top-5 \\
        \midrule
        \multicolumn{13}{l}{\cellcolor{lightgraybg}\textbf{External encoders}} \\
        \addlinespace[2pt]
        BrainIAC~\cite{tak2026:brainAIC} & 9.90$\pm0.08$ & 11.70$\pm0.43$  & 32.50$\pm0.87$ & 21.60$\pm0.00$ & 19.30$\pm0.00$ & 38.90$\pm0.00$ & 10.50$\pm0.01$ & 12.60$\pm0.27$ & 31.90$\pm0.53$ & 38.60$\pm0.00$ & \ul{18.80}$\pm0.00$ & 62.50$\pm0.00$ \\
        OpenMind~\cite{wald2025:openmind}& 10.50$\pm0.32$ & 9.60$\pm1.11$  & 34.30$\pm1.29$ & 17.60$\pm2.45$ & 15.00$\pm4.21$ & 43.90$\pm6.48$ & 10.80$\pm1.24$ & 12.30$\pm1.62$ & 39.70$\pm2.46$ & 32.70$\pm6.35$ & 12.50$\pm9.55$ & 56.20$\pm6.25$ \\
        VoCo~\cite{wu2026:voco}          & 11.40$\pm0.12$ & 18.90$\pm1.79$ & 49.90$\pm2.28$ & 19.30$\pm0.00$ & 13.20$\pm0.00$ & 25.40$\pm0.00$ & 16.30$\pm0.11$ & 25.70$\pm0.42$ & 50.30$\pm0.70$ & 32.40$\pm0.00$ & 6.20$\pm0.00$ & 68.80$\pm0.00$ \\
        3DINO (CLS)~\cite{3dinov2}       & 17.60$\pm0.06$ & 32.20$\pm0.51$ & 60.60$\pm0.61$ & 29.40$\pm0.00$ & 40.70$\pm0.00$ & 77.10$\pm0.00$ & 21.10$\pm0.09$ & 27.40$\pm1.21$ & 57.10$\pm0.59$ & 24.60$\pm0.00$ & 6.20$\pm0.00$ & 50.00$\pm0.00$ \\
        3DINO (GAP)~\cite{3dinov2}       & 20.10$\pm0.16$ & 36.90$\pm1.59$ & 64.50$\pm1.77$ & 30.20$\pm0.00$ & 35.70$\pm0.00$ & 76.40$\pm0.00$ & 28.00$\pm0.11$ & 44.60$\pm0.40$ & 77.60$\pm0.46$ & 28.10$\pm0.00$ & 6.20$\pm0.00$ & 56.20$\pm0.00$ \\
        SAM-Med3D~\cite{samed3d_tnnls}   & 13.61$\pm0.09$ & 21.01$\pm0.21$ & 51.59$\pm0.59$ & 18.30$\pm0.00$ & 15.36$\pm0.00$ & 36.43$\pm0.00$ & 15.32$\pm0.13$ & 24.48$\pm0.53$ & 45.75$\pm0.92$ & 31.00$\pm0.00$ & 6.25$\pm0.00$ & 68.75$\pm0.00$ \\
        Hi-End-MAE~\cite{Hi_End_MAE}     & 11.03$\pm0.07$ & 20.71$\pm0.70$ & 49.96$\pm2.19$ & 19.10$\pm0.00$ & 12.50$\pm0.00$ & 42.86$\pm0.00$ & 13.28$\pm0.05$ & 15.78$\pm0.27$ & 42.64$\pm1.02$ & 33.97$\pm0.00$ & 12.50$\pm0.00$ & 62.50$\pm0.00$ \\
        MambaMIM~\cite{MambaMIM}         & 10.78$\pm0.05$ & 13.54$\pm0.40$ & 36.58$\pm1.37$ & 18.27$\pm0.00$ & 13.93$\pm0.00$ & 38.21$\pm0.00$ & 13.44$\pm0.06$ & 12.72$\pm0.25$ & 25.14$\pm0.28$ & 36.37$\pm0.00$ & 18.75$\pm0.00$ & 50.00$\pm0.00$ \\
        MiM~\cite{MiM}                   & 9.00$\pm0.05$ & 9.56$\pm1.00$ & 31.49$\pm2.44$ & 18.26$\pm0.00$ & 15.36$\pm0.00$ & 42.50$\pm0.00$ & 9.58$\pm0.04$ & 11.37$\pm0.46$ & 28.32$\pm0.58$ & 33.97$\pm0.00$ & 12.50$\pm0.00$ & 62.50$\pm0.00$ \\
        GL-MAE~\cite{GL_MAE}             & 9.34$\pm0.04$ & 9.69$\pm0.12$ & 29.76$\pm0.97$ & 17.40$\pm0.00$ & 13.93$\pm0.00$ & 44.29$\pm0.00$ & 9.76$\pm0.01$ & 8.13$\pm0.27$  & 25.61$\pm0.82$ & 29.12$\pm0.00$ & 6.25$\pm0.00$  & 68.75$\pm0.00$ \\
        \midrule
        \multicolumn{13}{l}{\cellcolor{lightgraybg}\textbf{Controlled Med3DINO: CLS}} \\
        \addlinespace[2pt]
        Med3DINO-ISO      & 14.60$\pm0.13$ & 23.20$\pm1.02$ & 54.00$\pm1.35$ & 28.10$\pm0.00$ & 31.40$\pm0.00$ & 60.00$\pm0.00$ & 20.00$\pm0.12$ & 32.70$\pm0.22$ & 61.20$\pm0.53$ & 24.90$\pm0.00$ & 0.00 $\pm0.00$  &  62.50$\pm0.00$\\
        Med3DINO-REL      & 18.10$\pm0.06$ & 33.80$\pm0.77$ & 64.80$\pm0.24$ & 25.40$\pm0.00$ & 32.10$\pm0.00$ & 69.60$\pm0.00$ & 23.50$\pm0.10$ & 38.00$\pm0.66$ & 69.60$\pm0.16$ & 37.40$\pm0.00$ & 12.50$\pm0.00$  &  \ul{75.00}$\pm0.00$ \\
        Med3DINO-SA       & 19.90$\pm0.06$ & \ul{41.10}$\pm0.20$ & \ul{72.20}$\pm0.23$ & 31.10$\pm0.00$ & \ul{43.90}$\pm0.00$ & \ul{82.90}$\pm0.00$ & 28.30$\pm0.14$ & \ul{56.70}$\pm0.73$ & \ul{84.30}$\pm0.24$ & \textbf{48.00}$\pm0.00$ & \textbf{25.00}$\pm0.00$  &  \textbf{93.80}$\pm0.00$ \\
        \midrule
        \multicolumn{13}{l}{\cellcolor{lightgraybg}\textbf{Controlled Med3DINO: GAP}} \\
        \addlinespace[2pt]
        Med3DINO-ISO      & 16.20$\pm0.11$ & 26.80$\pm0.50$ & 54.60$\pm1.33$ & 25.90$\pm0.00$ & 26.10$\pm0.00$ & 56.10$\pm0.00$ & 21.60$\pm0.04$ & 27.80$\pm0.13$ & 59.60$\pm0.42$ & 33.60$\pm0.00$ & 12.50$\pm0.00$  &  62.50$\pm0.00$ \\
        Med3DINO-REL      & \ul{21.00}$\pm0.17$ & 38.50$\pm1.07$ & 65.80$\pm0.59$ & \ul{39.20}$\pm0.00$ & 43.20$\pm0.00$ & 72.50$\pm0.00$ & \ul{31.80}$\pm0.04$ & 54.00$\pm0.70$ & 78.00$\pm0.37$ & 35.50$\pm0.00$ & 12.50$\pm0.00$  &  68.80$\pm0.00$ \\
        Med3DINO-SA       & \textbf{27.20}$\pm0.24$ & \textbf{57.00}$\pm0.68$ & \textbf{78.10}$\pm0.42$ & \textbf{41.90}$\pm0.00$ & \textbf{48.60}$\pm0.00$ & \textbf{83.20}$\pm0.00$ & \textbf{44.80}$\pm0.09$ & \textbf{74.80}$\pm0.66$ & \textbf{91.20}$\pm0.05$ & \ul{42.00}$\pm0.00$ & \ul{18.80}$\pm0.00$  &  \ul{75.00}$\pm0.00$ \\
        \bottomrule
    \end{tabular}%
    }
    \vspace{-2.1em}
\end{table*}

\subsection{Anatomical Transfer Results}
\label{subsec:anatomical_generalization_results}
Anatomical Transfer evaluates whether organ identity remains recoverable from limited labeled support under case-disjoint evaluation. Table~\ref{tab:bench_generalization_cls} reports three linked quantities: $\mathrm{BA}_{centroid}$ for nearest-centroid separability, $\mathrm{BA}_{probe}$ for recoverability after limited adaptation, and $\Delta_{adapt}$ for the change relative to the centroid baseline. Within the controlled Med3DINO comparison, SA achieves the highest probe BA on TotalSegmentator CT--MR, MMWHS, and AMOS22 with both CLS and GAP features, with higher scores from GAP. Among the added encoders, Hi-End-MAE achieves the highest probe BA on all three datasets. Adaptation gain must be interpreted alongside final accuracy: a larger gain can reflect a weaker centroid baseline rather than a stronger final representation. The results therefore distinguish organ separability in frozen features from the recoverability achieved through limited supervision. Supplementary Sections~\ref{subsec:classifier_sensitivity} and~\ref{subsec:support_budget_sensitivity} examine classifier and support-budget dependence.
\begin{table*}[!t]
    \centering
    \scriptsize
    \setlength{\tabcolsep}{3pt}
    \caption{Case-disjoint organ classification with two support samples per organ and modality. Mean$\pm$SD are reported over three-seed evaluations. 3DINO and Med3DINO use CLS and GAP; other feature interfaces follow Table~\ref{tab:fm_comparison}.}
    \label{tab:bench_generalization_cls}
    \resizebox{\textwidth}{!}{%
    \begin{tabular}{@{}lrrr@{\hspace{8pt}}rrr@{\hspace{8pt}}rrr@{}}
        \toprule
        Method & \multicolumn{3}{c}{TotalSeg CT--MR~\cite{akinci:totalsegmentatorMR,wasserthal2023:totalsegmentatorCT}} & \multicolumn{3}{c}{MMWHS~\cite{zhuang2019:mmwhs_ct_mr}} & \multicolumn{3}{c}{AMOS22~\cite{ji2022:AMOS2022}} \\
        \cmidrule(lr){2-4} \cmidrule(lr){5-7} \cmidrule(lr){8-10}
         & BA$_{centroid}$ & BA$_{probe}$ & $\Delta_{adapt}$ & BA$_{centroid}$ & BA$_{probe}$ & $\Delta_{adapt}$ & BA$_{centroid}$ & BA$_{probe}$ & $\Delta_{adapt}$ \\
        \midrule
        \multicolumn{10}{l}{\cellcolor{lightgraybg}\textbf{External encoders}} \\
        \addlinespace[2pt]
        BrainIAC~\cite{tak2026:brainAIC}  & 17.00$\pm0.28$ & 20.40$\pm1.77$ & 3.40 $\pm1.51$ & 35.30$\pm2.70$ & 42.50$\pm2.82$ & 7.10 $\pm0.40$ & 16.70$\pm2.33$ & 21.60$\pm0.73$ & 4.90 $\pm2.84$\\
        OpenMind~\cite{wald2025:openmind} & 14.50$\pm2.94$ & 14.30$\pm3.07$ & -0.20$\pm0.13$ & 23.40$\pm3.67$ & 24.60$\pm2.93$ & 1.20 $\pm1.00$ & 18.80$\pm2.75$ & 18.50$\pm2.96$ & -0.20$\pm2.37$ \\
        VoCo~\cite{wu2026:voco}           & 25.80$\pm0.60$ & 36.10$\pm0.97$ & 10.20$\pm0.72$ & 29.20$\pm3.03$ & 47.60$\pm7.34$ & 18.40$\pm8.78$ & 37.10$\pm1.01$ & 52.70$\pm0.84$ & 15.60$\pm1.78$ \\
        3DINO (CLS)~\cite{3dinov2}        & 47.20$\pm2.24$ & 47.70$\pm2.19$ & 0.50 $\pm0.75$ & 57.40$\pm2.10$ & 61.60$\pm0.40$ & 4.20 $\pm1.82$ & 57.60$\pm2.66$ & 60.40$\pm1.64$ & 2.80 $\pm1.33$\\
        3DINO (GAP)~\cite{3dinov2}        & 47.00$\pm4.79$ & 52.50$\pm3.20$ & 5.50 $\pm2.30$ & 65.90$\pm5.51$ & 74.30$\pm4.24$ & 8.50 $\pm2.33$ & 64.40$\pm1.96$ & 70.10$\pm0.36$ & 5.70 $\pm1.65$\\
        SAM-Med3D~\cite{samed3d_tnnls}    & 34.56$\pm3.99$ & 36.54$\pm1.81$ & 1.98$\pm2.18$  & 34.79$\pm1.88$  & 55.29$\pm8.41$  & 20.50$\pm9.22$ & 35.68$\pm0.27$ & 43.90$\pm3.92$ & 8.22$\pm4.15$ \\
        Hi-End-MAE~\cite{Hi_End_MAE}      & 31.80$\pm1.28$ & 44.39$\pm1.76$ & 12.60$\pm3.01$ & 44.05$\pm10.80$ & 61.11$\pm17.66$ & 17.06$\pm7.15$ & 40.74$\pm2.84$ & 57.98$\pm1.66$ & 17.25$\pm4.43$ \\
        MambaMIM~\cite{MambaMIM}          & 22.73$\pm3.04$ & 32.79$\pm2.42$ & 10.05$\pm5.47$ & 37.70$\pm12.40$ & 47.09$\pm8.66$  & 9.39$\pm5.67$  & 24.50$\pm4.31$ & 43.95$\pm2.93$ & 19.45$\pm3.17$ \\
        MiM~\cite{MiM}                    & 14.83$\pm2.92$ & 19.15$\pm0.78$ & 4.32$\pm2.94$  & 23.81$\pm3.83$  & 46.16$\pm12.41$ & 22.35$\pm8.73$ & 15.67$\pm0.62$ & 33.26$\pm5.22$ & 17.59$\pm5.36$ \\
        GL-MAE~\cite{GL_MAE}              & 14.35$\pm1.09$ & 18.34$\pm2.88$ & 3.98$\pm2.55$  & 33.33$\pm14.59$ & 42.33$\pm12.02$ & 8.99$\pm4.42$  & 16.57$\pm2.31$ & 24.50$\pm3.82$ & 7.93$\pm1.77$ \\
        \midrule
        \multicolumn{10}{l}{\cellcolor{lightgraybg}\textbf{Controlled Med3DINO: CLS}} \\
        \addlinespace[2pt]
        Med3DINO-ISO            & 43.10$\pm2.66$ & 43.10$\pm1.80$ & 0.00$\pm0.95$ & 43.10$\pm8.36$ & 53.30$\pm7.20$ & 10.20$\pm2.00$ & 51.70$\pm0.95$ & 55.10$\pm1.17$ & 3.40$\pm1.40$ \\
        Med3DINO-REL            & 45.80$\pm3.04$ & 46.10$\pm3.09$ & 0.30$\pm0.10$ & 50.40$\pm6.97$ & 58.10$\pm5.57$ & 7.70$\pm1.39$  & 58.20$\pm3.65$ & 63.00$\pm3.42$ & 4.80$\pm0.25$ \\
        Med3DINO-SA             & 48.20$\pm4.80$ & 48.60$\pm3.39$ & 0.40$\pm2.31$ & 57.30$\pm5.50$ & 62.80$\pm4.80$ & 5.50$\pm1.28$  & 60.80$\pm3.32$ & 64.90$\pm2.81$ & 4.10$\pm1.72$ \\
        \midrule
        \multicolumn{10}{l}{\cellcolor{lightgraybg}\textbf{Controlled Med3DINO: GAP}} \\
        \addlinespace[2pt]
        Med3DINO-ISO            & 37.50$\pm2.09$ & 44.30$\pm4.17$ & 6.80$\pm2.14$ & 54.20$\pm6.35$ & 67.10$\pm4.96$ & 12.80$\pm3.99$ & 55.90$\pm2.78$ & 61.50$\pm5.44$ & 5.60$\pm3.46$ \\
        Med3DINO-REL            & \ul{52.30}$\pm1.81$ & \ul{54.80}$\pm1.97$ & 2.50$\pm1.33$ & \textbf{75.10}$\pm1.00$ & \ul{79.00}$\pm3.08$ & 3.80$\pm2.58$  & \ul{67.40}$\pm2.14$ & \ul{71.10}$\pm3.38$ & 3.70$\pm1.66$ \\
        Med3DINO-SA             & \textbf{55.30}$\pm1.57$ & \textbf{58.80}$\pm1.75$ & 3.50$\pm0.76$ & \ul{74.50}$\pm2.33$ & \textbf{80.00}$\pm0.83$ & 5.60$\pm1.60$  & \textbf{71.70}$\pm3.12$ & \textbf{76.20}$\pm2.39$ & 4.50$\pm0.91$ \\
        \bottomrule
    \end{tabular}%
    }
    % \vspace{-0.8em}
\end{table*}

\begin{figure*}[!t]
    \centering
    \includegraphics[width=\textwidth]{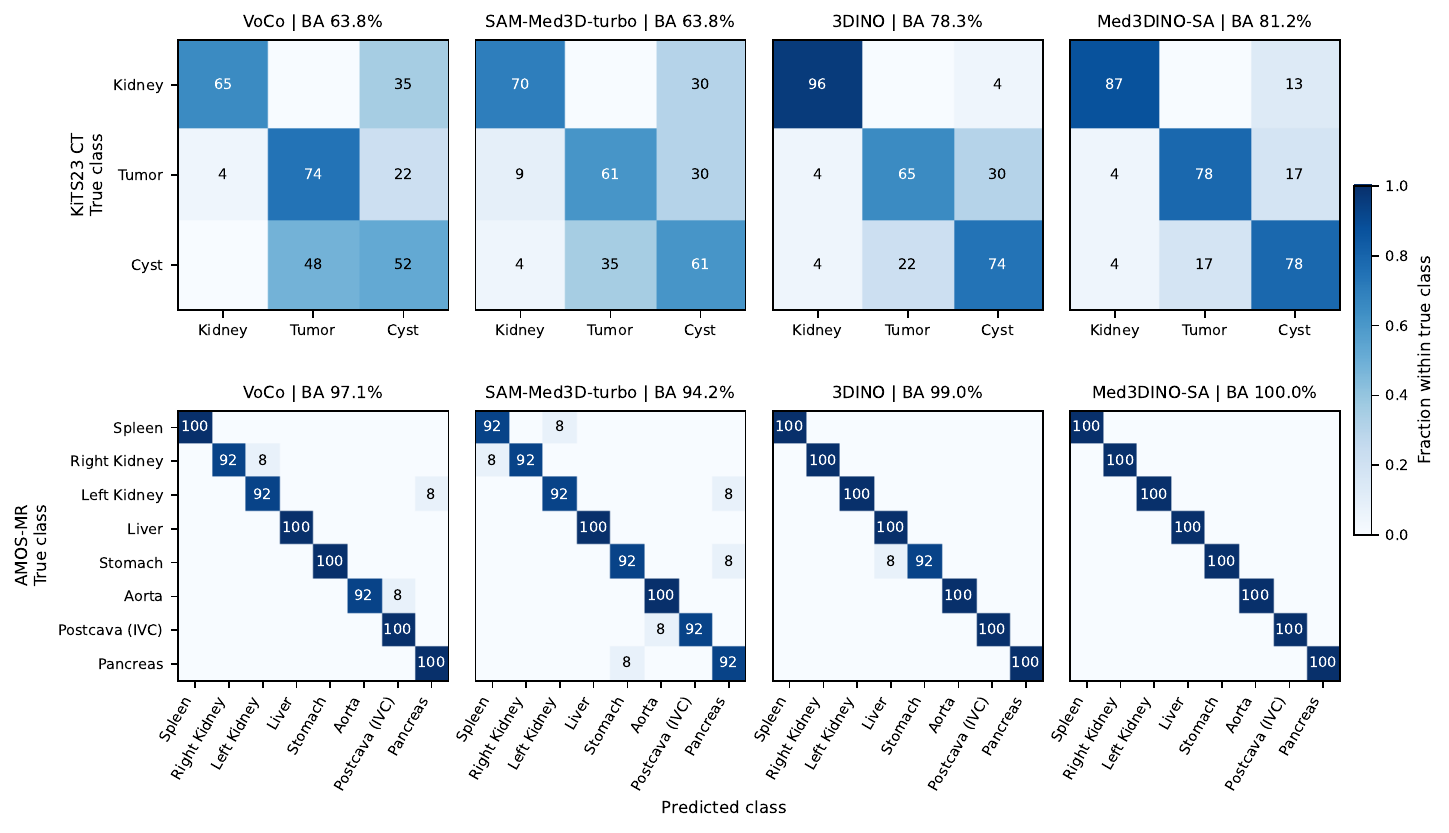}
    \caption{Case-disjoint organ classification of frozen crop features on KiTS23 CT (top) and AMOS-MR (bottom). Logistic classifiers use 40 and 38 labeled, context-inclusive crops per class, respectively. Cells show row-normalized percentages; panel titles report BA. 3DINO and Med3DINO-SA use CLS features, VoCo uses multiscale GAP, and SAM-Med3D-turbo uses post-neck GAP.}
    \label{fig:heldout_crop_classification}
    \vspace{-1.5em}
\end{figure*}

An additional case-disjoint linear-probe evaluation with larger labeled support (Fig.~\ref{fig:heldout_crop_classification}) reveals class-specific differences beyond aggregate BA. All four encoders achieve high organ recall on AMOS-MR, whereas KiTS23 errors predominantly involve tumor--cyst confusion. Med3DINO-SA achieves the highest BA on both cohorts, although 3DINO has higher kidney recall, showing that overall recoverability does not imply uniform class-level superiority.

In summary, within the controlled Med3DINO comparison, REL has lower reported CLS drift than SA on the four displayed acquisition datasets, whereas SA achieves the highest reported cross-modal retrieval scores and probe BA on the three multi-organ CT--MR datasets with both CLS and GAP. Semantic transfer and CKA remain dataset- and feature-dependent. The retrieval and probe results span both mixed-exposure and pretraining-unseen datasets. REFORM-3D therefore separates embedding stability from anatomical recoverability rather than establishing one global model ranking.
\begin{table*}[!t]
    \centering
    \scriptsize
    \caption{Segmentation performance across 10 downstream datasets, grouped by their relationship to Med3DINO pretraining: seen/unseen datasets and unseen-modality transfer. Exposure labels refer to Med3DINO and do not automatically apply to external encoders. Dice (\%) and HD95 (mm) are reported; higher Dice and lower HD95 are better. TS-CT denotes TotalSegmentator-CT~\cite{wasserthal2023:totalsegmentatorCT}.}
    \label{tab:segmentation-results}
    \setlength{\tabcolsep}{3pt}
    \resizebox{\textwidth}{!}{%
    \begin{tabular}{l*{10}{cc}}
        \toprule
        Method
        & \multicolumn{10}{c}{Seen datasets} & \multicolumn{8}{c}{Unseen datasets} & \multicolumn{2}{c}{Unseen modality} \\
        \cmidrule(lr){2-11} \cmidrule(lr){12-19} \cmidrule(lr){20-21}
        & \multicolumn{2}{c}{BTCV~\cite{landman2015:BTCV}}
        & \multicolumn{2}{c}{BraTS-GLI~\cite{BraTS21}}
        & \multicolumn{2}{c}{ISLES22~\cite{hernandez:ISLES22}}
        & \multicolumn{2}{c}{WORD~\cite{luo2022:WORD}}
        & \multicolumn{2}{c}{TS-CT~\cite{wasserthal2023:totalsegmentatorCT}}
        & \multicolumn{2}{c}{AMOS-CT~\cite{ji2022:AMOS2022}}
        & \multicolumn{2}{c}{AMOS-MR~\cite{ji2022:AMOS2022}}
        & \multicolumn{2}{c}{KiTS23~\cite{heller2023kits21}}
        & \multicolumn{2}{c}{LA-SEG~\cite{LA-Seg}}
        & \multicolumn{2}{c}{TDSC-ABUS~\cite{luo2025:TDSC_ABUS}} \\ \cmidrule(lr){2-3} \cmidrule(lr){4-5} \cmidrule(lr){6-7} \cmidrule(lr){8-9}
        \cmidrule(lr){10-11} \cmidrule(lr){12-13} \cmidrule(lr){14-15} \cmidrule(lr){16-17} \cmidrule(lr){18-19} \cmidrule(lr){20-21}
        & Dice & HD95 & Dice & HD95 & Dice & HD95 & Dice & HD95 & Dice & HD95 & Dice & HD95 & Dice & HD95 & Dice & HD95 & Dice & HD95 & Dice & HD95 \\
        % \midrule
        % nnUNet~\cite{isensee18:nnUNet}    & 73.30 & 43.81 &  &  & 82.50 & --  &  &  &  &  & 84.81 & -- &  &  & 69.15  & --  &  &  &  &  \\
        \midrule
        MedDINOv3~\cite{li2025:meddinov3} & 78.79  & -- & -- & -- & -- & -- & -- & -- & -- & -- & 87.38 & -- & -- & -- & 70.68 & --  & -- & -- & -- & -- \\
        3DINO~\cite{3dinov2}              & \textbf{82.10} & 9.4  & 91.80 & 3.90 & 63.39 & 11.08& 85.46 & 6.83 & 86.78 & 5.83 & 85.56 & 5.01 & 80.69 & 5.54 & 80.75 & 13.87 & 92.40 & 5.20 & 47.70 & 89.10  \\
        % 3DINO~\cite{3dinov2}$^\dagger$& 85.56 & 5.01 & 80.69 & 5.54 & 82.10 & 9.4 & 91.80 & 3.90 & 63.39 & 11.08 & 80.75 & 13.87 & 85.46 & 6.83 & 92.40 & 5.20 & 86.78 & 5.83 & 47.70 & 89.10  \\
        \midrule
        Med3DINO-ISO                      & 64.71 & 15.56& 91.70 & 3.46 & 64.98 & 9.87 & 86.10 & 7.50 & 86.94 & 4.96 & 83.35 & 4.99 & 82.12 & \textbf{4.69} & 88.00 & 19.61& 95.69 & 1.96 & 72.78 & 75.61  \\
        Med3DINO-REL                      & 65.50 & 7.94 & 91.77 & 3.54 & 67.76 & 9.37 & 86.65 & 6.30 & 88.79 & 3.83 & 85.46 & 4.28 & 82.07 & 5.58          & 88.11 & 14.57& 95.17 & 1.97 & 74.94  & \underline{68.72}  \\
        Med3DINO-SA                       & 75.69 & \textbf{5.76} & \textbf{92.17} & \textbf{3.13} & \textbf{70.70} & \textbf{6.42} & \textbf{86.79} & \textbf{6.07} & \textbf{88.96} & \textbf{3.52} & \textbf{87.88} & \textbf{3.82} & \textbf{82.17} & 4.83 & \textbf{88.50} & \textbf{9.15} & \textbf{95.80} & \textbf{1.90} & \textbf{75.91} & \textbf{65.29}  \\        
        \bottomrule
    \end{tabular}
    }
    \vspace{-1.95em}
\end{table*}

\vspace{-0.5em}
\subsection{Downstream Performance}
\label{subsec:downstream_performance_results}

Downstream results are included as secondary evidence rather than as the basis of the framework's main claims. Table~\ref{tab:segmentation-results} gives the segmentation summary, organized into seen datasets, unseen datasets, and one unseen-modality transfer setting. Within Med3DINO family, SA achieves the highest reported Dice on all ten datasets and the lowest HD95 on nine; ISO achieves lower HD95 on AMOS-MR. Among external comparisons, 3DINO achieves higher BTCV Dice. These results are consistent with the cross-modal retrieval and probe results, but downstream scores also depend on decoder choice, supervision, and optimization. We therefore use downstream performance to show practical transfer relevance on the studied tasks, not to replace the representation-level benchmark. The segmentation protocol is detailed in Supplementary Section~\ref{subsec:downstream_segmentation_protocol}.

% \section{Limitations}
% \label{sec:limitations}
% REFORM-3D characterizes frozen representations on specific acquisition, cross-modal retrieval, and case-disjoint anatomical-transfer surfaces, but does not establish broader invariance or replace downstream evaluation. Acquisition robustness is evaluated primarily through controlled resampling, which does not reproduce the full variability of clinical image acquisition and should therefore be interpreted as geometric stress testing rather than complete acquisition realism. The REL–SA findings are also established within a single architecture and self-supervised objective; their generality requires matched geometric-policy comparisons under other pretraining recipes. 

\section{Conclusion}
\label{sec:conclusion}
We presented REFORM-3D, a representation-centric framework for understanding frozen 3D medical representations under acquisition variation, cross-modal anatomy, and case-disjoint anatomical transfer. Instead of treating downstream task performance as sufficient evidence of representation quality, the framework evaluates whether features remain stable when spacing changes while anatomy is held fixed, whether labeled structures remain recoverable across modalities under balanced bidirectional retrieval, and whether organ identity is recoverable from limited labeled support under case-disjoint evaluation. Across Med3DINO and external encoders, these evaluations reveal distinct strengths in representation stability and semantic or anatomical recoverability that downstream task scores alone do not distinguish. By making these properties measurable under controlled evaluation settings, the benchmark clarifies semantic and anatomical recoverability from frozen 3D representations, alongside their sensitivity to acquisition geometry and modality differences.
% By making these properties measurable under controlled evaluation settings, the benchmark provides a more precise account of what 3D medical foundation models encode and where their transfer behavior is anatomically grounded or remains sensitive to acquisition and modality shortcuts.

\textbf{Limitations.} REFORM-3D characterizes frozen representations on specific acquisition, cross-modal retrieval, and case-disjoint anatomical transfer surfaces, but does not establish broader invariance or replace downstream evaluation. Acquisition robustness is evaluated primarily through controlled resampling, which does not reproduce the full variability of clinical image acquisition and should therefore be interpreted as geometric stress testing rather than complete acquisition realism. The REL–SA findings are also established within a single architecture and self-supervised objective; their generality requires matched geometric-policy comparisons under other pretraining recipes. 

\textbf{Societal Impact.} REFORM-3D can support more transparent evaluation and informed selection of medical image encoders. Overreliance on benchmark rankings may obscure performance differences across patient groups and acquisition settings. Downstream clinical applications require validation in their intended settings, assessment across relevant patient groups, and safeguards for patient privacy.

\begin{ack}
This research was funded by the Center for Autonomous Robotic Systems (CARS), Khalifa University of Science and Technology (KU-CARS).
\end{ack}

\newpage

\bibliographystyle{unsrt_short}
\bibliography{refs}

%%%%%%%%%%%%%%%%%%%%%%%%%%%%%%%%%%%%%%%%%%%%%%%%%%%%%%%%%%%%%%%%%%%%%%%%%%%%
\newpage
\appendix
%%%%%%%%%%%%%%%%%%%%%%%%%%%%%%%%%%%%%%%%%%%%%%%%%%%%%%%%%%%%%%%%%%%%%%%%%%%%

%%%%%%%%%%%%%%%%%%%%%%%%%%%%%%%%%%%%%%%%%%%%%%%%%%%%%%%%%%%%%%%%%%%%%%%%%%%%
\begin{center}
{\Large \textbf{REFORM-3D: A Representation-Centric Evaluation Framework}}\\
\vspace{0.5em}
{\Large \textbf{for 3D Medical Vision Foundation Models}}\\
\vspace{0.7em}
{\large Supplementary Material}
\end{center}
\vspace{1em}
%%%%%%%%%%%%%%%%%%%%%%%%%%%%%%%%%%%%%%%%%%%%%%%%%%%%%%%%%%%%%%%%%%%%%%%%%%%%
The supplementary material documents the data and protocols underlying REFORM-3D and the controlled Med3DINO comparison, together with additional analyses and reproducibility resources.

\begin{description}
    \setlength{\itemsep}{2pt}
    \setlength{\parskip}{0pt}
    \item[Section~\ref{sec:pretraining_details}.] Pretraining Details and Data Composition
    \item[Section~\ref{sec:benchmark_dataset_construction}.] Benchmark Dataset Construction
    \item[Section~\ref{sec:benchmark_protocol_metric}.] Benchmark Protocol and Metric Definitions
    \item[Section~\ref{sec:additional_results_and_extended_tables}.] Additional Benchmarking Results and Extended Tables
    \item[Section~\ref{sec:implementation_reproducibility_benchmarking}.] Implementation and Reproducibility
\end{description}
%%%%%%%%%%%%%%%%%%%%%%%%%%%%%%%%%%%%%%%%%

\section{Pretraining Details and Data Composition}
\label{sec:pretraining_details}
The Med3DINO encoders are pretrained with a volumetric adaptation of DINOv2~\cite{oquab2024:dinov2} using 3D Vision Transformers and student-teacher self-supervised learning, following 3DINO~\cite{3dinov2}. This section presents the data and training details needed to interpret the benchmark. The main benchmark evaluates frozen encoders, so the pretraining description is included to clarify the data scale, modality breadth, and geometric assumptions compared by the ISO, REL, and SA regimes.

\subsection{Corpus Composition and Source Inventory}
\label{subsec:dataset_breakdown}
\begin{figure}[!t]
  \centering
  % Image on the left
  \begin{minipage}[c]{0.55\linewidth}
    \centering
    \includegraphics[width=\linewidth]{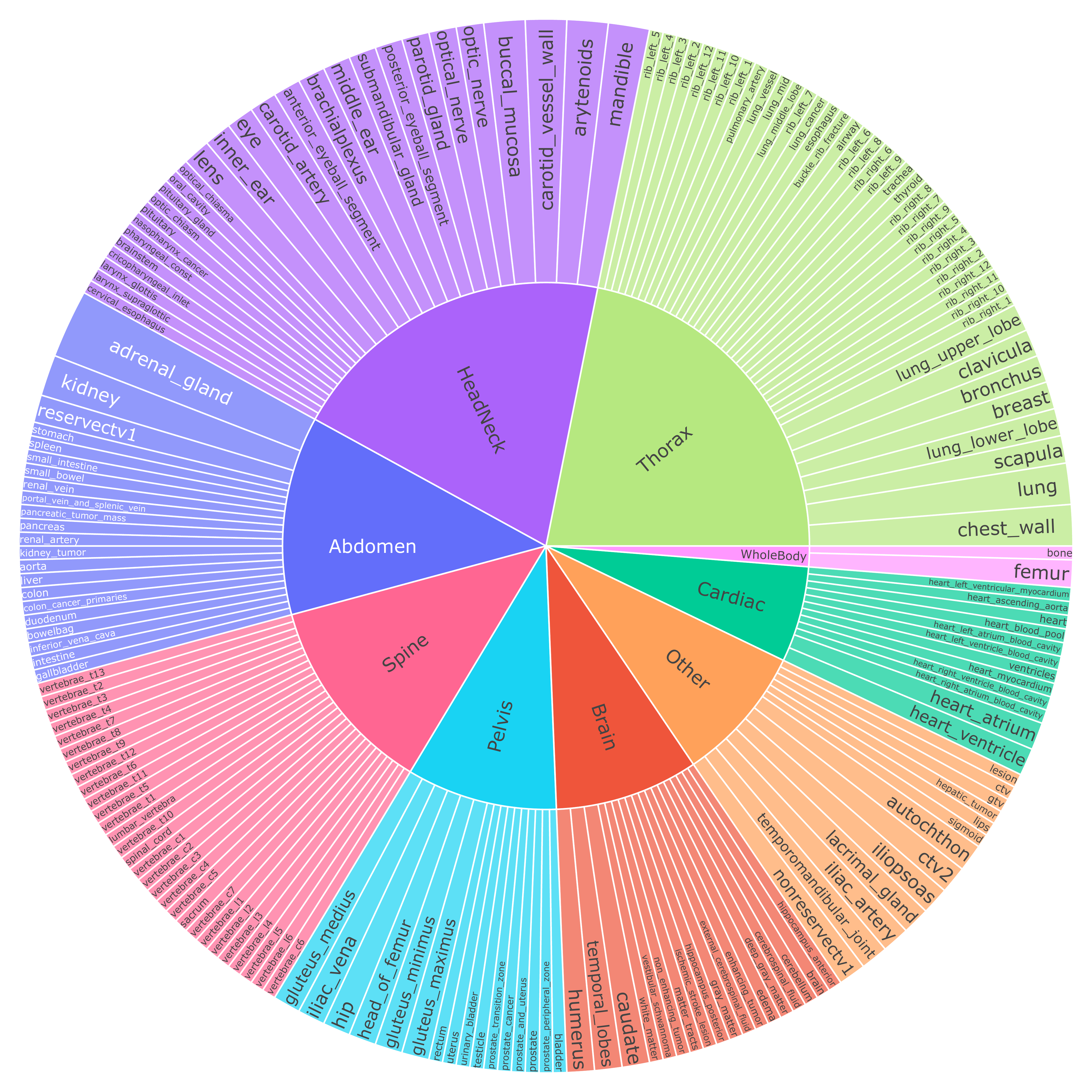}
  \end{minipage}%
  \hfill
  % Caption on the right
  \begin{minipage}[c]{0.40\linewidth}
    \caption{Summary of anatomical coverage in the pretraining corpus. The listed structures are representative labels from the corpus mapping rather than an exhaustive benchmark label set.}
    \label{fig:organ_coverage_summary}
  \end{minipage}
\end{figure}

The pretraining corpus comprises 62,638 unlabeled 3D multimodal volumes retained after quality curation of the source collections in Table~\ref{tab:pretraining_corpus_list}. Individual modality and sequence images are treated as separate volumes during pretraining. The inventory lists 64,880 dataset-level samples across 65 source datasets. These counts sum entries across source collections rather than unique patients or individual image volumes. The validation and preprocessing stages are described in Section~\ref{subsec:pretraining_preprocessing}. CT and MR dominate the retained corpus, while PET, CTA, and CBCT extend the acquisition spectrum. Table~\ref{tab:dataset_summary_compact} summarizes the retained corpus by modality and lists representative pathology-associated labels. Figure~\ref{fig:organ_coverage_summary} illustrates anatomical coverage from the corpus mapping of 238 fine-grained labels consolidated into 182 organ cores across 9 coarse anatomical regions. This taxonomy describes pretraining exposure, not the label set of every benchmark evaluation.
\begin{table}[ht]
    \centering
    \small
    \setlength{\tabcolsep}{4pt}

    \begin{minipage}[t]{0.48\linewidth}
        \centering
        \begin{tabular}{lrr}
            \toprule
            \textbf{Modality} & \textbf{Volumes} & \textbf{Share (\%)} \\
            \midrule
            CT   & 34,052 & 54.4 \\
            MR   & 23,245 & 37.1 \\
            PET  & 3,368  & 5.4  \\
            CTA  & 1,070  & 1.7  \\
            CBCT & 903    & 1.4  \\
            \midrule
            Total Corpus & 62,638 & 100\% \\
            \bottomrule
        \end{tabular}
        \vspace{0.3em}
    \end{minipage}
    % \hfill
    \begin{minipage}[t]{0.48\linewidth}
        \centering
        \begin{tabular}{lp{0.6\linewidth}}
            \toprule
            \textbf{Region} & \textbf{Pathology-associated labels} \\
            \midrule
            Abdomen &
            colon cancer primaries; kidney tumor; liver tumor; pancreatic tumor \\
            Brain &
            enhancing tumor; ischemic stroke lesion; non-enhancing tumor; vestibular schwannoma \\
            Thorax &
            buckle rib fracture; lung cancer \\
            Head and neck &
            nasopharynx cancer \\
            Pelvis &
            prostate cancer \\
            Other &
            lesion, unspecified \\
            \bottomrule
        \end{tabular}

        \vspace{0.3em}
    \end{minipage}
    \caption{Pretraining corpus composition. Left: modality distribution over 62,638 unlabeled 3D volumes. Right: organ and lesion labels associated with disease findings in the corpus mapping.}
    \label{tab:dataset_summary_compact}
    \vspace{-2.2em}
\end{table}

\begin{table}[!t]
    \centering
    \scriptsize
    \setlength{\tabcolsep}{3pt}
    \renewcommand{\arraystretch}{0.95}
        \caption{Source-collection inventory, totaling 64,880 listed samples. Counts follow dataset-level units rather than individual image volumes: a paired CT/PET sample contributes two volumes, while multi-sequence MR cases contribute separate sequence images. The listed samples account for at least 81,319 source image volumes before curation, including documented modality and sequence expansions. Access requirements are separate from licenses; ``No explicit'' denotes an unspecified license in this inventory, not unrestricted use.}
        \label{tab:pretraining_corpus_list}
    \newsavebox{\sourceinventorybox}
    \begin{lrbox}{\sourceinventorybox}
    \begin{tabular}{@{}>{\raggedright\arraybackslash}p{0.28\textwidth}cc>{\raggedright\arraybackslash}p{0.21\textwidth}lr@{}}
        \toprule
        Dataset & Mod. & Version & License & Access & \shortstack{Source samples} \\
        \midrule
        AbdomenCT-1K~\cite{ma2021:abdomenct_1k} & CT & V1 & CC BY 4.0 & Request & 1062 \\
        ATM22~\cite{ZHANG2023102957} & CT & V1 & CC BY-NC & Agreement & 500 \\
        BraTS15~\cite{6975210} & MR & V1 & No explicit & Registration & 384 \\
        BraTS18~\cite{BraTS2018} & MR & V1 & MIT & Custom (BraTS terms) & 542 \\
        BraTS19~\cite{Bakas2017TCGAGlioma} & MR & V1 & No explicit & Registration & 626 \\
        BraTS20~\cite{Bakas2017TCGAGlioma} & MR & V1 & No explicit & Request & 369 \\
        BraTS21~\cite{BraTS21} & MR & V1 & CC BY 4.0 & Registration & 2040 \\
        BrainPTM~\cite{nelkenbaum2020automatic} & MR & V1 & Custom & Agreement & 75 \\
        BTCV\_Cervix~\cite{landman2015:BTCV} & CT & V1 & CC BY 4.0 & Registration & 30 \\
        COSMOS22~\cite{cosmos2022} & MR & V1 & No explicit & Request & 50 \\
        COVID19CTscans~\cite{COVID-19} & CT & V1 & CC BY-NC-SA & Open & 20 \\
        CrossMoDA21~\cite{shapey2021segmentation} & MR & V1 & CC BY-SA 4.0 & Open & 379 \\
        CrossMoDA22~\cite{DORENT2023:CrossMoDA} & MR & V2 & CC BY-NC-SA 4.0 & Open & 755 \\
        CT-ORG~\cite{CT-ORG} & CT & V1 & CC BY 3.0 & Open & 140 \\
        CTPelvic1k~\cite{liu2021deep} & CT & V1 & CC BY-NC-SA 4.0 & Open & 1184 \\
        FLARE21~\cite{flare21} & CT & V1 & CC BY-NC-SA 4.0 & Registration & 511 \\
        FLARE22~\cite{FLARE22-LDH2024} & CT & V2 & No explicit & Registration & 2300 \\
        ISLES22~\cite{hernandez:ISLES22} & MR & V1 & CC BY 4.0 & Registration & 250 \\
        ISLES24~\cite{riedel2025isles24} & CT/MR & V1 & CC BY-NC-SA 4.0 & Registration & 149 \\
        KIPA22~\cite{HE2021:KIPA} & CT & V1 & CC BY-NC-ND 4.0 & Agreement & 130 \\
        JHU-Stroke~\cite{liu2023:jhu_stroke} & MR & V6 & CC BY & Custom (DUA) & 2888 \\
        LAScarQS22~\cite{AtrialGeneral} & MR & V1 & CC BY-NC-ND 4.0 & Agreement & 194 \\
        LiTS~\cite{bilic2019liver} & CT & V1 & CC BY-NC-SA & Registration & 201 \\
        LUNA16~\cite{setio2017validation} & CT & V1 & CC BY 4.0 & Open & 888 \\
        MSD BrainTumour~\cite{antonelli2022medical} & MR & V1 & CC BY-SA 4.0 & Open & 750 \\
        MSD Colon~\cite{antonelli2022medical} & CT & V1 & CC BY-SA 4.0 & Open & 190 \\
        MSD Heart~\cite{antonelli2022medical} & MR & V1 & CC BY-SA 4.0 & Open & 30 \\
        MSD HepaticVessel~\cite{antonelli2022medical} & CT & V1 & CC BY-SA 4.0 & Open & 443 \\
        MSD Hippocampus~\cite{antonelli2022medical} & MR & V1 & CC BY-SA 4.0 & Open & 394 \\
        MSD Liver~\cite{antonelli2022medical} & CT & V1 & CC BY-SA 4.0 & Open & 201 \\
        MSD Lung~\cite{antonelli2022medical} & CT & V1 & CC BY-SA 4.0 & Open & 96 \\
        MSD Pancreas~\cite{antonelli2022medical} & CT & V1 & CC BY-SA 4.0 & Open & 420 \\
        MSD Prostate~\cite{antonelli2022medical} & MR & V1 & CC BY-SA 4.0 & Open & 48 \\
        MSD Spleen~\cite{antonelli2022medical} & CT & V1 & CC BY-SA 4.0 & Open & 61 \\
        AbdomenAtlas~\cite{li2024:SuPreM} & CT & V1 & CC BY-NC-SA 4.0 & Open & 9262 \\
        PICAI~\cite{SahaBosmaTwilt2024} & MR & V2 & CC BY-NC 4.0 & Registration & 1500 \\
        Promise09~\cite{miccai2009_prostate_dataset} & MR & V1 & CC BY-SA & No explicit & 15 \\
        Promise12~\cite{LITJENS:Promise12} & MR & V2 & No explicit & Registration & 100 \\
        RibFrac2020~\cite{ribfracchallenge2024} & CT & V1 & CC BY-NC 4.0 & Open & 660 \\
        SLIVER07~\cite{heimann2009comparison} & CT & V3 & No explicit & Open & 30 \\
        STACOM-SLAWT~\cite{karim2018algorithms} & CT/MR & V1 & CC BY 4.0 & Open & 20 \\
        StructSeg19~\cite{structseg2019} & CT & V1 & No explicit & Open & 100 \\
        Totalsegmentator-CT~\cite{wasserthal2023:totalsegmentatorCT} & CT & V1 & CC BY 4.0 & Open & 1204 \\
        VerSe19~\cite{sekuboyina2021verse} & CT & V1 & CC BY-SA 4.0 & Open & 160 \\
        VerSe20~\cite{liebl2021verse} & CT & V2 & CC BY-SA 4.0 & Open & 319 \\
        WORD~\cite{luo2022:WORD} & CT & V0.1 & GPL-3.0 & Open & 150 \\
        RSNA-2023~\cite{rudie2024ratic} & CT & V1 & Custom & Registration & 4274 \\
        RAD-Chest-CT~\cite{DRAELOS2021:RAD-Chest-CT} & CT & V1 & CC BY-NC-ND 4.0 & Request & 3630 \\
        ULS-2023~\cite{DEGRAUW2025:ULS23} & CT & V2 & CC BY-NC-SA 4.0 & Open & 6514 \\HECKTOR-2025 task 1~\cite{hecktor2025} & CT/PET & V1 & CC BY-NC-SA 4.0 & Request & 680 \\
        HECKTOR-2025 task 2~\cite{hecktor2025} & CT/PET & V1 & CC BY-NC-SA 4.0 & Request & 666 \\
        autoPET (FDG)~\cite{gatidis2022:autoPET_v1} & CT/PET & I & CC BY-NC 4.0 & Registration & 1014 \\
        autoPET (FDG+PSMA)~\cite{jeblick2026:autoPET_v2} & CT/PET & III & CC BY-NC 4.0 & Registration & 1548 \\
        Duke Liver~\cite{Duke_Liver} & MR & V2 & CC BY-NC-ND 4.0 & Request & 2146 \\
        ImageCAS~\cite{zeng2023:imagecas} & CTA & V1 & Apache-2.0 & Open & 1000 \\
        ImageTAAD~\cite{song2025:ImageTAAD} & CTA & V1 & Apache-2.0 & Open & 120 \\
        ImageTBAD~\cite{yao2021imagetbad} & CTA & V1 & Apache-2.0 & Open & 100 \\
        CTSpine1K~\cite{CTSpine1K} & CT & V1 & CC BY-NC-SA 4.0 & Open & 1005 \\
        fastMRI (Brain)~\cite{zbontar2018fastMRI} & MR & V1 & Custom & Request & 6970 \\
        % fastMRI (Breast)~\cite{solomon2025fastmri_breast} & MR & V1 & Custom & Request & 300 \\
        % fastMRI (Prostate)~\cite{tibrewala2023fastmri_prostate} & MR & V1 & Custom & Request & 312 \\
        % fastMRI (Knee)~\cite{zbontar2018fastMRI} & MR & V1 & Custom & Request & 1398 \\
        fastMRI (Breast)~\cite{zbontar2018fastMRI} & MR & V1 & Custom & Request & 300 \\
        fastMRI (Prostate)~\cite{zbontar2018fastMRI} & MR & V1 & Custom & Request & 312 \\
        fastMRI (Knee)~\cite{zbontar2018fastMRI} & MR & V1 & Custom & Request & 1398 \\
        ToothFairy~\cite{ToothFairy} & CBCT & V1 & CC BY-SA 4.0 & Registration & 532 \\
        ToothFairy2~\cite{ToothFairy2} & CBCT & V1 & CC BY-SA 4.0 & Registration & 480 \\
        STS-Tooth~\cite{wang2025:STS_Tooth}   & CBCT & V1 & CC BY 4.0 & Open & 371 \\
        % \midrule
        % \multicolumn{5}{r}{Total source samples} & \textbf{63,680} \\
        \midrule
        \multicolumn{5}{l}{\textbf{Subtotals:} CT: 35,685;\ MR: 22,515;\ CT/MR: 169;\ CT/PET: 3,908;\ CTA: 1,220;\ CBCT: 1,383} & \textbf{Total source samples: 64,880} \\
        \bottomrule
    \end{tabular}
    \end{lrbox}
    \ifdim\wd\sourceinventorybox>\textwidth
        \sbox{\sourceinventorybox}{\resizebox{\textwidth}{!}{\usebox{\sourceinventorybox}}}
    \fi
    \ifdim\dimexpr\ht\sourceinventorybox+\dp\sourceinventorybox\relax>0.80\textheight
        \resizebox*{!}{0.80\textheight}{\usebox{\sourceinventorybox}}
    \else
        \usebox{\sourceinventorybox}
    \fi
    \vspace{-2.5em}
\end{table}

\subsection{Field-of-View and Spacing Distributions}
\label{app:fov_spacing_distributions}
To quantify acquisition geometry in the pretraining corpus, we paired each volume's voxel spacing $(s_1,s_2,s_3)$ with its image shape $(n_1,n_2,n_3)$ and computed physical field of view (FOV) along axis $a$ as $\mathrm{FOV}_a = n_a s_a$. The summaries average spacing and FOV over the first two recorded image axes and use the third axis separately, denoted in-plane and through-plane, respectively. Median spacing is $1.30$ mm in-plane and $2.00$ mm through-plane, with $1^{\mathrm{st}}$--$99^{\mathrm{th}}$ percentile ranges of $0.36$--$5.47$ mm and $0.50$--$6.60$ mm, respectively. The corresponding percentile FOV ranges are $140.00$--$814.57$ mm in-plane and $103.00$--$1005.00$ mm through-plane.

\begin{figure}[t]
    \centering
    \includegraphics[width=0.48\linewidth]{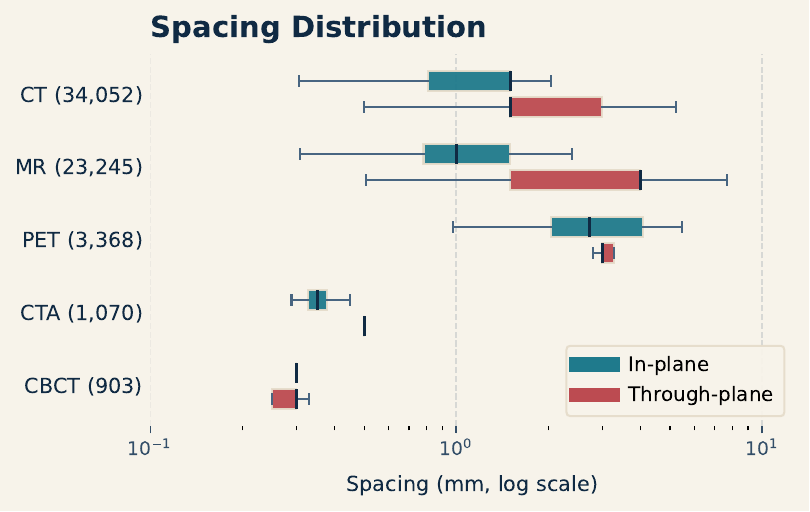}\hfill
    \includegraphics[width=0.48\linewidth]{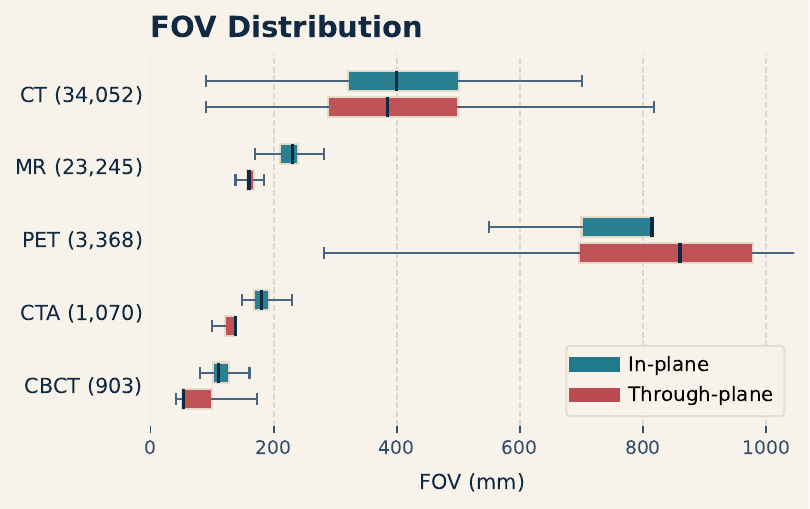}
    \caption{Pretraining-corpus acquisition geometry. Left: voxel-spacing distributions. Right: physical field-of-view distributions. These figures show why fixed voxel-space crops can correspond to different physical anatomical support across the corpus.}
    \label{fig:pretraining_geometry_distributions}
    \vspace{-1.0em}
\end{figure}

Figure~\ref{fig:pretraining_geometry_distributions} summarizes variation in voxel spacing and physical coverage. Consequently, a crop specified only in voxel units does not correspond to fixed physical anatomical support across scans. This motivates the ISO, REL, and SA comparison, but corpus geometry alone does not establish representation robustness; acquisition sensitivity is evaluated through the controlled same-scan spacing perturbations defined in Section~\ref{sec:benchmark_protocol_metric}.

\subsection{Preprocessing and Validation}
\label{subsec:pretraining_preprocessing}
Corpus curation combined spatial screening, file-integrity validation, and preprocessing-compatibility checks. Within the selected modalities, volumes were required to contain at least 32 slices and 96 voxels along both in-plane axes. Validation assessed compressed-stream and NIfTI integrity, consistency between headers and voxel arrays, spacing and affine validity, physical coverage, finite voxel values, and intensity variation. Modality-specific intensity-encoding correction and clipping preceded checks of compatibility with normalization, foreground cropping, and FOV handling. Volumes failing validation were excluded, with rejection reasons recorded.

During loading, a single time point or channel was selected when necessary, and \texttt{NaNs} were replaced by the sample-level mean. Intensities were normalized using the $0.05^{\mathrm{th}}$ and $99.95^{\mathrm{th}}$ percentiles and mapped to $[-1,1]$. Volumes were then cropped to foreground defined by intensities above $-1$, with voxel-spacing metadata used to identify and reorder the slice axis.

Before multi-crop augmentation, offline validation assessed foreground content and intensity variation in the normalized, foreground-cropped volume. Voxels with intensity above $-0.95$ had to occupy at least $5\%$ of the cropped volume; the image intensity variance had to be at least $0.01$, and a nonempty foreground bounding box was required. The final pretraining manifest contains 62,638 volumes that passed all validation stages, including these foreground and intensity-variance checks. Foreground selection and validation were intensity-based and did not require segmentation labels.

\subsection{Self-Supervised Training Objective}
\label{subsec:self_supervised_training_objective}

Med3DINO adapts 3DINO~\cite{3dinov2} to volumetric medical imaging with a 3D Vision Transformer backbone. The default training backbone is ViT-L/16 with cubic $16^3$ patch embedding. For a $96^3$ global crop, the encoder produces a $6\times6\times6$ patch grid, corresponding to 216 patch tokens, plus one CLS token. The self-supervised objective combines DINO self-distillation on CLS tokens, iBOT masked patch prediction on patch tokens, and KoLeo regularization on student CLS features. The teacher network is an exponential moving average of the student and is not updated by backpropagation.

The default configuration uses two global crops of size $96^3$ and eight local crops of size $48^3$. DINO and iBOT projection heads each use 65,536 prototypes, a bottleneck dimension of $256$, $3$ layers, and hidden dimension $2048$. iBOT masks patch tokens with mask probability $0.5$ and mask-ratio range $[0.2,0.75]$; the default configuration uses a separate iBOT head. KoLeo loss is applied with weight $0.1$. Teacher temperature is $0.07$ after warmup, warmup teacher temperature is $0.04$, and teacher momentum increases from $0.992$ to $1.0$. All three variants use AdamW with a per-GPU batch size of $64$ for $100$ epochs of $1,250$ iterations each ($125,000$ updates), base learning rate $0.002$, $10$ warmup epochs, weight decay from $0.04$ to $0.4$, gradient clipping at $3.0$, patch-embedding learning-rate multiplier $0.2$, and layerwise decay $0.9$. The base learning rate is scaled as $0.002\sqrt{B/1024}$, where $B$ is the global batch size.

\subsection{Geometric Pretraining Regimes}
\label{subsec:geometric_pretraining_regimes}
The three regimes use the same corpus and shared training setup, while varying input geometry, crop definition, and rotation handling. ISO resamples volumes to $1.5\times1.5\times1.5$ mm spacing before voxel-relative cropping; REL retains native spacing and uses voxel-relative crops; SA retains native spacing and combines relative in-plane with physical through-plane sampling. The comparison is interpreted at the level of frozen representations learned under these complete geometric policies, not as an isolated ablation of a single augmentation parameter. Table~\ref{tab:geometric_policy_controls} separates the shared settings from the varying components.

In the relative regime, global and local crops are sampled with 3D random resized crops. Global crops use in-slice scale range $[0.48,1.0]$ and cross-slice scale range $[0.5,1.0]$; local crops use in-slice scale range $[0.16,0.48]$ and cross-slice scale range $[0.2,0.5]$. Rotations are applied without an anisotropy gate. In the spacing-aware regime, crops combine relative in-plane sampling with physical through-plane sampling. Global crops use through-plane physical scale range $[120,280]$ mm and local crops use $[40,100]$ mm. Rotations are gated by anisotropy with threshold $2.0$ so that cross-plane rotations are avoided when they would create implausible orientation changes for highly anisotropic data.
\begin{table}[t]
    \centering
    \scriptsize
    \caption{Fixed and varying components of the Med3DINO geometric-policy comparison. The intervention is the complete geometric policy, including input geometry, crop definition, and rotation handling; it is not an isolated change to crop coordinates.}
    \label{tab:geometric_policy_controls}
    \begin{tabular}{@{}lp{0.72\textwidth}@{}}
        \toprule
        Component & Controlled setting \\
        \midrule
        Shared data and model & 62,638-volume corpus; initialization policy; ViT-L/16. \\
        Shared optimization & DINO/iBOT/KoLeo; non-geometric augmentations; crop counts and output sizes; batch size; AdamW schedule; 125K updates. \\
        ISO geometry & Resampling to $1.5\times1.5\times1.5$ mm followed by voxel-relative crops. \\
        REL geometry & Retained anisotropy with voxel-relative crops. \\
        SA geometry & Retained anisotropy with physical through-plane crop extents and anisotropy-aware rotation handling. \\
        \bottomrule
    \end{tabular}
    \vspace{-2.2em}
\end{table}

\section{Benchmark Dataset Construction}
\label{sec:benchmark_dataset_construction}
Each module uses a dataset construction rule specific to its representation question. Acquisition Robustness compares resampled variants of the same scan. Cross-Modal Anatomical Alignment requires a shared label definition and sufficient samples in both modalities. Anatomical Transfer uses organ-level support and query samples from disjoint cases. Dataset exposure refers specifically to the Med3DINO pretraining corpus, not to external encoders pretraining.

\subsection{Acquisition Robustness Datasets}
\label{subsec:acquisition_dataset_construction}

The controlled Acquisition Robustness setting starts from source scans and creates spacing variants of the same scan. Table~\ref{tab:benchmark_cohort_sizes} summarizes full and sampled
manifest sizes from the source scans used for controlled evaluation. For CT and CTA settings, the chosen target spacings are $(1.0,1.0,1.0)$ mm, $(1.0,1.0,3.0)$ mm, and $(1.0,1.0,5.0)$ mm. These variants represent isotropic, moderately anisotropic, and highly anisotropic through-plane sampling while holding in-plane spacing fixed. For MR settings, the controlled protocol keeps each scan's native in-plane spacing and varies target through-plane spacing using $2.0, 4.0,$ and $6.0$ mm targets. This MR protocol avoids forcing MR scans onto a CT-like in-plane spacing while still testing through-plane sensitivity.

The same-scan construction is the anchor evidence: comparisons are made among variants derived from the same source image. Native-spacing analyses are built separately by assigning original scans to anisotropy bins: \(r = \frac{\max(s_x,s_y,s_z)}{\min(s_x,s_y,s_z)}.\)
% \begin{equation}
% r = \frac{\max(s_x,s_y,s_z)}{\min(s_x,s_y,s_z)}.
% \end{equation}
The bins are near-isotropic $[1.0,1.5)$, moderately anisotropic $[1.5,3.0)$, and highly anisotropic $[3.0,\infty)$. Because native-spacing bins contain different scans, they are interpreted only as observational support and not as the primary robustness claim. The controlled cohorts are smaller because source selection is constrained by native anisotropy: CT/CTA evaluation uses the available near-isotropic scans, while MR evaluation uses moderately anisotropic scans due to limited near-isotropic availability, summarized in Table~\ref{tab:benchmark_cohort_sizes}.

\subsection{Cross-Modal Anatomical Alignment Datasets}
\label{subsec:cross_modal_dataset_construction}
TotalSegmentator CT--MR, MMWHS, and AMOS22 provide 12, 7, and 13 shared organ classes, respectively. Organ records identify the source case, modality, anatomical label, image, and mask. The primary extraction crops the original organ bounding box with a five-voxel margin and retains surrounding image content; it therefore evaluates context-assisted organ retrieval, not purely organ-intrinsic features. Query and target pools are balanced by organ within each direction.

CHAOS uses eight queries and eight targets per direction under identifier-based liver pairing between different CT and MR patients (Section~\ref{subsec:cross_modal_metric_definitions}); it is not pooled with multi-organ retrieval. The CT--PET extension uses HECKTOR-2025 primary-tumor and metastatic-lymph-node categories and excludes same-patient targets (Section~\ref{subsec:ct_pet_extension}). These surfaces test retrieval under their declared label and modality definitions, not voxel-level registration or general modality invariance.
\begin{table}[!t]
    \centering
    \scriptsize
    \setlength{\tabcolsep}{5pt}
    \caption{Cohort sizes for the benchmark evaluations. Acquisition counts distinguish full manifests, sampled manifests, and source scans used for controlled resampling; each source scan contributes three spacing variants. Retrieval entries report query/target (Q/T) organ or lesion-category records per direction.}
    \label{tab:benchmark_cohort_sizes}
    \begin{tabular}{@{}lrrr@{}}
        \toprule
        \multicolumn{4}{l}{\textit{Acquisition Robustness}} \\
        Dataset & Full-manifest scans & Sampled scans & Controlled scans \\
        \midrule
        AbdomenAtlas~\cite{li2024:SuPreM}         & 9,208 & 3,000 & 1,000 \\
        KiTS23~\cite{heller2023kits21}                  & 489 & 489 & 69 \\
        ImageCAS~\cite{zeng2023:imagecas}               & 1,000 & 1,000 & 708 \\
        TotalSegmentator-MR~\cite{akinci:totalsegmentatorMR} & 616 & 616 & 131 \\
        JHU-Stroke~\cite{liu2023:jhu_stroke}            & 1,853 & 1,138 & 131 \\
        \midrule
        \multicolumn{4}{l}{\textit{Multi-organ CT--MR retrieval}} \\
        Dataset & Source cases: CT/MR & CT$\rightarrow$MR: Q/T & MR$\rightarrow$CT: Q/T \\
        \midrule
        TotalSegmentator~\cite{wasserthal2023:totalsegmentatorCT,akinci:totalsegmentatorMR} & 1,204/616 & 6,648/1,032 & 1,032/6,648 \\
        MMWHS~\cite{zhuang2019:mmwhs_ct_mr}                                                 & 20/20 & 140/140 & 140/140 \\
        AMOS22~\cite{ji2022:AMOS2022}                                                       & 300/60 & 3,666/715 & 715/3,666 \\
        \midrule
        \multicolumn{4}{l}{\textit{Lesion-category CT--PET retrieval}} \\
        Dataset & Source cases: CT/PET
        & CT$\rightarrow$PET: Q/T & PET$\rightarrow$CT: Q/T \\
        \midrule
        HECKTOR-2025~\cite{hecktor2025}  & 678/678 & 200/200 & 200/200 \\
        \bottomrule
    \end{tabular}
    \vspace{-1.8em}
\end{table}

\subsection{Anatomical Transfer Datasets}
\label{subsec:anatomical_generalization_dataset_construction}
Anatomical Transfer uses TotalSegmentator CT--MR, MMWHS, and AMOS22. Every evaluated organ, including the target organ, receives two support samples per modality. Support cases are selected before constructing the query pool; all records from any support case are excluded from queries, across organs and modalities. Queries are then balanced by organ and modality. This is multiclass organ classification with both modalities represented in support and queries.

The reported cohorts contain 84, 18, and 53 query samples per organ and modality for TotalSegmentator, MMWHS, and AMOS22, respectively. With 12, 7, and 13 classes, these correspond to 48/2,016, 28/252, and 52/1,378 total support/query records. These support/query counts are consistent across evaluated encoders and feature interfaces. Case-disjoint evaluation does not establish that organ classes were absent from pretraining.

% \subsection{Label Harmonization and Eligibility}
% \label{subsec:label_harmonization_eligibility}
% Manifests specify sample identifiers, modality, canonical anatomical labels, image and mask paths, and source-case identifiers. Eligibility is determined before sampling from available features. Original organ masks must satisfy the minimum size of 100 voxels used for region extraction. Multi-organ retrieval requires at least two classes supported in both modalities; organ transfer additionally requires case-disjoint support and query records for every class. The CHAOS identifier-based pairing protocol is evaluated separately rather than applying this multi-organ eligibility rule to a single label.

% The protocol-stability experiments retain a fixed eligible cohort across interventions and reject missing or duplicate sample IDs, incomplete feature bundles, and stale preprocessing signatures. These safeguards apply to the protocol-stability experiments. The physical-mask and organ-label controls are defined in Section~\ref{subsec:protocol_stability}.

\section{Benchmark Protocol and Metric Definitions}
\label{sec:benchmark_protocol_metric}
All benchmark encoders remain frozen, including when semantic-transfer classifiers or linear probes are fitted. Classifier training updates only the classifier, not the encoder. External comparisons use the same metric definitions but retain the checkpoint-specific input preprocessing and feature interfaces listed in Table~\ref{tab:fm_comparison}; causal attribution to geometric pretraining is restricted to ISO/REL/SA triplets.

\subsection{Frozen Feature Extraction}
\label{subsec:frozen_feature_extraction}
For 3DINO and Med3DINO, CLS is the final normalized class token, and GAP is the mean of final normalized patch tokens over spatial positions, preserving the feature channels. The resulting CLS and GAP vectors are then $\ell_2$-normalized. Med3DINO evaluations use the registered $96^3$ crop checkpoints; other encoders retain their declared input and feature interfaces. BrainIAC uses its first patch token; OpenMind and MiM use global average pooling; VoCo pools multiscale features; SAM-Med3D pools post-neck features; MambaMIM pools its final-stage features; Hi-End-MAE and GL-MAE use CLS. These are frozen representations, not task-specific segmentation outputs.

Crop caches are shared only when their image preprocessing is identical. Embeddings remain separate by checkpoint, feature type, and intervention. Repeated evaluation reuses deterministic features when the cohort and preprocessing are unchanged; changing the evaluation seed does not require another encoder forward pass.

\subsection{Acquisition Robustness Metrics}
\label{subsec:acquisition_metric_definitions}

For a scan $x_i$, a frozen encoder $f_\theta$, and a deterministic preprocessing and resampling transform $T(x_i;s)$ for target spacing $s$, the controlled representation is
\begin{equation}
z_i(s)=f_\theta(T(x_i;s)).
\end{equation}
The controlled setting compares $z_i(s)$ across spacing variants of the same scan.

\paragraph{Drift.}
For target spacings $a$ and $b$, let $z_i^{(a)}=z_i(a)$ and $z_i^{(b)}=z_i(b)$. Their cosine distance defines within-scan drift:
\begin{equation}
d_i(a,b)=1-\cos\left(z_i^{(a)},z_i^{(b)}\right).
\end{equation}
Lower drift means that the embedding of the same scan moves less under spacing perturbation. Drift is an individual-scan stability metric and does not by itself show whether the cohort-level feature geometry is preserved.

\paragraph{CKA.}
Centered kernel alignment (CKA)~\cite{kornblith2019:CKA} is computed between feature matrices from two spacing variants over the same controlled sample set. Let $X,Y\in\mathbb{R}^{n\times d}$ denote the centered feature matrices for the same $n$ scans under two spacing variants, with rows corresponding to scans and columns to feature dimensions. Linear CKA follows
\begin{equation}
\mathrm{CKA}(X,Y)=\frac{\lVert Y^\top X \rVert_F^2}{\lVert X^\top X \rVert_F \lVert Y^\top Y \rVert_F},
\end{equation}
where $\lVert\cdot\rVert_F$ denotes the Frobenius norm. Higher CKA indicates stronger preservation of cohort-level feature organization across spacing variants. Drift and CKA therefore provide complementary information: the former measures changes in individual embedding directions, whereas the latter measures preservation of centered cohort-level feature organization. 

\paragraph{Semantic Transfer.}
For each source spacing $a$ and target spacing $b$, classifiers are trained on frozen source-spacing features and tested on target-spacing features from held-out cases. Cases, semantic targets, five-fold splits, and classifier configuration are fixed across spacing comparisons. Standardization is fitted only on the source training fold and applied unchanged to the target test fold. Each eligible binary target uses logistic regression with \texttt{lbfgs}, $C=1$, and at most 1,000 iterations, without warm-starting or condition-specific tuning. Folds use stratification when the target structure and class counts permit it, and otherwise use seeded shuffled splits. 
A binary target is evaluated in a fold only when the source training set and target test set each contain at least two positive and two negative examples.

Dataset-specific targets are defined in Table~\ref{tab:semantic_targets}. For each directed spacing transfer, balanced accuracy is averaged over eligible targets within each fold and then across evaluated folds. The headline score averages the six directed transfers among the three spacing variants. It is not a common semantic score that can be pooled across datasets. The classifier-sensitivity analysis varies regularization and replaces logistic regression with nearest-centroid classification (Section~\ref{subsec:classifier_sensitivity}). These comparisons characterize recoverability under controlled resampling, not spacing effects independent of image interpolation or scanner physics.

\begin{table}[t]
    \centering
    \scriptsize
    \caption{Acquisition semantic-target definitions. Targets are derived from the original annotations and held fixed across spacing variants. Constant or unsupported targets are filtered before evaluation. Foreground-volume tertiles use foreground voxel counts in the source masks, not a common anatomical label vocabulary.}
    \label{tab:semantic_targets}
    \begin{tabular}{@{}lp{0.72\textwidth}@{}}
        \toprule
        Dataset & Target definition \\
        \midrule
        AbdomenAtlas & Binary presence of eligible structures from the annotation-label mapping. \\
        KiTS23 & Kidney, tumor, and cyst presence candidates; only cyst presence is informative in the stored controlled cohort. \\
        ImageCAS & One-versus-rest indicators for coronary-mask foreground-volume tertiles. \\
        TS-MR & Structure presence from the individual segmentation masks. \\
        JHU-Stroke & One-versus-rest indicators for lesion-mask foreground-volume tertiles. \\
        \bottomrule
    \end{tabular}
    \vspace{-2.0em}
\end{table}

\subsection{Cross-Modal Retrieval Metrics}
\label{subsec:cross_modal_metric_definitions}
For each query, candidates from the other modality are ranked by cosine similarity. Multi-organ pools contain equal target counts per organ, and positives share the query's organ label. Let $C(q)$ denote the candidate pool, $P(q)$ its positive subset, and $c_k$ the candidate at rank $k$. Then
\begin{equation}
\mathrm{AP}(q)=\frac{1}{|P(q)|}\sum_{k=1}^{|C(q)|}\mathrm{Prec}@k(q)\,\mathbf{1}[c_k\in P(q)],
\qquad
\mathrm{mAP}_{a\rightarrow b}=\frac{1}{|Q_a|}\sum_{q\in Q_a}\mathrm{AP}(q).
\end{equation}
Here, $a$ and $b$ denote source and target modalities, $Q_a$ is the query set from modality $a$, and $\mathrm{Prec}@k(q)$ is the fraction of positives among the first $k$ ranked candidates. The bidirectional score is $\mathrm{mAP}_{a\leftrightarrow b}=(\mathrm{mAP}_{a\rightarrow b}+\mathrm{mAP}_{b\rightarrow a})/2$. Top-1 and Top-5 measure whether any positive occurs within the first one or five candidates and use the same equal-direction averaging. CHAOS uses identifier-based liver pairing between different CT and MR patients; HECKTOR defines positives by lesion category after excluding same-patient targets. These evaluation settings are reported separately.

\subsection{Anatomical Transfer Metrics}
\label{subsec:anatomical_generalization_metric_definitions}
For each seed, support and query sets are balanced by class and modality and case-disjoint. The standardizer is fitted on support features only. Nearest-centroid classification assigns each standardized query to the closest support-class mean by Euclidean distance. The linear probe uses logistic regression with \texttt{lbfgs}, $C=1$, and at most 2,000 iterations on the same standardized support set. Neither classifier changes the encoder.
\begin{equation}
\mathrm{BA}=\frac{1}{|\mathcal{O}|}\sum_{o\in\mathcal{O}}\mathrm{Recall}(o),
\qquad
\Delta_{\mathrm{adapt}}=\mathrm{BA}_{\mathrm{probe}}-\mathrm{BA}_{\mathrm{centroid}}.
\end{equation}
Each class recall is measured on case-disjoint queries.
% , with the target class included in support at the same budget as other classes. 
Adaptation gain is computed within each seed before aggregation. Negative gains indicate that the fitted probe performs worse than nearest centroid on that evaluation split. This protocol evaluates limited-label organ classification. %, not zero-shot transfer to anatomy absent from pretraining.

\subsection{Uncertainty and Paired Comparisons}
\label{subsec:uncertainty_protocol}
Results report mean$\pm$sample SD over evaluation seeds 42, 123, and 456 for fixed checkpoints; scores and SDs are scaled by 100, and accuracy/retrieval differences are percentage points. Paired estimates and intervals refer to their corresponding cohorts and runs. REL--ISO, SA--ISO, and SA--REL use 50,000 bootstrap resamples (seed 2026). Acquisition and retrieval hierarchically resample seeds and paired units: scans for drift, scan rows with recomputed CKA, six directed spacing-transfer summaries for semantic transfer, and queries preserving equal retrieval-direction weights. Organ-classification intervals cross seeds and organ recalls. Primary intervals are centered and symmetric; classifier-sensitivity intervals are percentile-based. Bonferroni adjustment across the three contrasts within each dataset--feature--metric setting, and classifier where applicable, targets 95\% family-wise coverage. Intervals containing zero establish neither separation nor equivalence. Uncertainty reflects evaluation randomness, not independent pretraining runs, and does not establish superiority over external encoders without corresponding paired tests.

\section{Additional Benchmarking Results and Extended Tables}
\label{sec:additional_results_and_extended_tables}
The main paper reports the primary acquisition, cross-modal retrieval, and organ transfer comparisons. Here, we provide the additional JHU-Stroke acquisition results and paired geometric-policy comparisons, followed by sensitivity analyses, organ- and direction-level breakdowns, spacing-severity comparisons, pretraining-unseen evaluation, and a CT--PET retrieval extension. Each analysis is interpreted within its stated dataset, feature interface, and protocol, using the metric and uncertainty definitions in Section~\ref{sec:benchmark_protocol_metric}.

\subsection{Acquisition Robustness Extensions}
\label{subsec:acquisition_extended_results}
The supplementary acquisition results comprise JHU-Stroke performance in Table~\ref{tab:bench_acquisition_controlled_seeded_full} and paired ISO/REL/SA comparisons across all five acquisition datasets in Table~\ref{tab:acquisition_paired_analysis}. The JHU-Stroke table complements the four-dataset main comparison and retains all evaluated encoders, including both CLS and GAP for 3DINO and Med3DINO. Within each evaluation run, drift averages the two reference-to-target comparisons, CKA averages the off-diagonal spacing comparisons, and semantic transfer averages the six directed source-to-target transfers. Semantic-transfer scores remain dataset-specific because their targets differ. Together, these analyses test whether geometric stability and semantic recoverability lead to the same policy comparisons across datasets and feature interfaces.
\begin{table}[!t]
    \centering
    \small
    \setlength{\tabcolsep}{6pt}
    \caption{Controlled same-scan acquisition robustness on JHU-Stroke. Entries are mean$\pm$SD over evaluation seeds (42,123,456) for fixed checkpoints; 0.00 may reflect rounding. Semantic transfer uses lesion-mask foreground-volume tertiles. External comparisons are contextual; geometric-policy attribution is restricted to ISO/REL/SA.}
    \label{tab:bench_acquisition_controlled_seeded_full}
    \begin{tabular}{@{}lrrr@{}}
        \toprule
        Method & \multicolumn{3}{c}{JHU-Stroke~\cite{liu2023:jhu_stroke}} \\
        \cmidrule(lr){2-4}
        & Drift$\downarrow$ & CKA$\uparrow$ & Semantic Transfer$\uparrow$ \\
        \midrule
        \multicolumn{4}{l}{\cellcolor{lightgraybg}\textbf{External encoders}} \\
        BrainIAC~\cite{tak2026:brainAIC}    & 1.30$\pm0.00$ & \ul{99.80}$\pm0.00$ & 59.90$\pm2.14$ \\
        OpenMind~\cite{wald2025:openmind}   & \ul{0.00}$\pm0.00$ & 97.70$\pm0.42$ & 54.41$\pm0.55$ \\
        VoCo~\cite{wu2026:voco}             & 0.10$\pm0.00$ & 85.72$\pm0.00$ & 66.40$\pm1.72$ \\
        3DINO~\cite{3dinov2} (CLS)          & 1.43$\pm0.00$ & 95.30$\pm0.00$ & \ul{70.73}$\pm0.96$ \\
        3DINO~\cite{3dinov2} (GAP)          & 0.10$\pm0.00$ & 98.45$\pm0.00$ & 66.26$\pm2.44$ \\
        SAM-Med3D~\cite{samed3d_tnnls}      & 0.05$\pm0.00$ & 98.10$\pm0.00$ & 63.60$\pm2.19$ \\
        Hi-End-MAE~\cite{Hi_End_MAE}        & 0.09$\pm0.00$ & 98.89$\pm0.00$ & 60.57$\pm1.90$ \\
        MambaMIM~\cite{MambaMIM}            & 0.04$\pm0.00$ & 98.98$\pm0.00$ & 58.50$\pm0.52$ \\
        MiM~\cite{MiM}                      & 0.00$\pm0.00$ & 99.75$\pm0.00$ & 69.57$\pm0.98$ \\
        GL-MAE~\cite{GL_MAE}                & \textbf{0.00}$\pm0.00$ & \textbf{99.83}$\pm0.00$ & 65.56$\pm2.27$ \\
        \midrule
        \multicolumn{4}{l}{\cellcolor{lightgraybg}\textbf{Controlled Med3DINO: CLS}} \\
        Med3DINO-ISO    & 4.30$\pm0.00$ & 87.20$\pm0.00$ & 65.30$\pm0.95$ \\
        Med3DINO-REL    & 1.16$\pm0.00$ & 96.11$\pm0.00$ & 69.94$\pm1.78$ \\
        Med3DINO-SA     & 3.30$\pm0.00$ & 92.73$\pm0.00$ & \textbf{74.52}$\pm1.36$ \\
        \midrule
        \multicolumn{4}{l}{\cellcolor{lightgraybg}\textbf{Controlled Med3DINO: GAP}} \\
        Med3DINO-ISO    & 0.21$\pm0.00$ & 94.02$\pm0.00$ & 66.45$\pm0.66$ \\
        Med3DINO-REL    & 0.14$\pm0.00$ & 99.18$\pm0.00$ & 68.41$\pm0.72$ \\
        Med3DINO-SA     & 0.33$\pm0.00$ & 97.46$\pm0.00$ & 70.71$\pm0.89$ \\
        \bottomrule
    \end{tabular}
    \vspace{-0.5em}
\end{table}

On JHU-Stroke, REL has the lowest mean drift and highest mean CKA within both Med3DINO feature families, whereas SA has the highest mean semantic-transfer score. Near-zero reported drift among external encoders likewise does not identify the highest transfer score. These descriptive results motivate joint assessment of embedding stability and target recoverability; they do not establish pairwise significance across encoders.

For CLS features, the paired SA--REL comparisons favor REL for drift on four datasets, with no statistically supported difference on TS-MR. CKA favors REL on AbdomenAtlas, KiTS23, and JHU-Stroke, but SA on TS-MR; the ImageCAS interval includes zero. The paired semantic-transfer comparisons favor SA on AbdomenAtlas, KiTS23, and TS-MR, and REL on ImageCAS, with no statistically supported SA--REL difference on JHU-Stroke. Relative to ISO, SA has supported transfer advantages on AbdomenAtlas, TS-MR, and JHU-Stroke, with no statistically supported difference on KiTS23 or ImageCAS; REL improves transfer on all datasets except KiTS23, where ISO is favored.

Overall, these comparisons show that a policy can better stabilize embeddings without yielding better semantic transfer. Acquisition robustness therefore requires complementary assessment of within-scan drift, cohort-level geometry, and recoverability of the specified targets; no single metric supports a uniform policy ranking.
\begin{table*}[!t]
    \centering
    \scriptsize
    \setlength{\tabcolsep}{4pt}
    \caption{Paired ISO/REL/SA acquisition comparisons across five datasets. Entries are signed effects [Bonferroni-adjusted confidence intervals], scaled by 100. For $A$ vs.\ $B$, effects are $B-A$ for drift and $A-B$ for CKA and transfer; positive values favor $A$, whereas negative values favor B. Intervals follow Section~\ref{subsec:uncertainty_protocol} (50,000 bootstrap resamples; bootstrap seed 2026), with adjustment across the three contrasts within each dataset--feature--metric setting targeting 95\% family-wise coverage. Bold entries exclude zero before rounding; intervals containing zero do not establish equivalence. Uncertainty reflects evaluation of fixed checkpoints, not independent pretraining runs. 
    %TS-MR denotes TotalSegmentator-MR.
    }
    \label{tab:acquisition_paired_analysis}
    \resizebox{\textwidth}{!}{%
    \begin{tabular}{@{}lrrrrr@{}}
        \toprule
        Comparison & AbdomenAtlas~\cite{li2024:SuPreM} & KiTS23~\cite{heller2023kits21} & ImageCAS~\cite{zeng2023:imagecas} & TotalSegmentator-MR~\cite{akinci:totalsegmentatorMR} & JHU-Stroke~\cite{liu2023:jhu_stroke} \\
        \midrule
        \multicolumn{6}{l}{\cellcolor{lightgraybg}\textbf{Drift$\downarrow$ --- CLS}} \\
        REL vs.\ ISO & \textbf{+7.31 [6.96, 7.65]} & \textbf{+4.00 [3.03, 4.98]} & \textbf{-3.81 [-4.16, -3.47]} & +0.33 [-0.19, 0.85] & \textbf{+1.45 [1.26, 1.64]} \\
        SA vs.\ ISO & +0.09 [-0.25, 0.42] & \textbf{-4.22 [-5.26, -3.17]} & \textbf{-18.45 [-18.90, -18.01]} & +0.28 [-0.08, 0.65] & \textbf{-3.43 [-3.69, -3.18]} \\
        SA vs.\ REL & \textbf{-7.22 [-7.41, -7.02]} & \textbf{-8.22 [-8.83, -7.61]} & \textbf{-14.64 [-15.09, -14.20]} & -0.05 [-0.54, 0.44] & \textbf{-4.88 [-5.05, -4.72]} \\
        \midrule
        \multicolumn{6}{l}{\cellcolor{lightgraybg}\textbf{CKA$\uparrow$ --- CLS}} \\
        REL vs.\ ISO & \textbf{+1.61 [1.39, 1.82]} & -0.21 [-0.54, 0.13] & \textbf{-10.36 [-11.11, -9.61]} & -0.04 [-0.19, 0.11] & \textbf{+0.68 [0.58, 0.78]} \\
        SA vs.\ ISO & +0.21 [-0.05, 0.46] & \textbf{-0.99 [-1.35, -0.64]} & \textbf{-10.43 [-11.08, -9.77]} & \textbf{+0.34 [0.18, 0.51]} & \textbf{-3.08 [-3.57, -2.58]} \\
        SA vs.\ REL & \textbf{-1.40 [-1.59, -1.21]} & \textbf{-0.79 [-1.03, -0.54]} & -0.06 [-0.93, 0.81] & \textbf{+0.38 [0.24, 0.52]} & \textbf{-3.75 [-4.25, -3.26]} \\
        \midrule
        \multicolumn{6}{l}{\cellcolor{lightgraybg}\textbf{Semantic transfer$\uparrow$ --- CLS}} \\
        REL vs.\ ISO & \textbf{+3.35 [2.10, 4.59]} & \textbf{-7.80 [-10.78, -4.82]} & \textbf{+1.83 [1.03, 2.63]} & \textbf{+14.79 [13.42, 16.15]} & \textbf{+5.21 [3.56, 6.86]} \\
        SA vs.\ ISO & \textbf{+7.06 [6.00, 8.12]} & +3.58 [-1.34, 8.51] & +0.50 [-0.66, 1.67] & \textbf{+19.01 [17.55, 20.47]} & \textbf{+3.50 [1.78, 5.22]} \\
        SA vs.\ REL & \textbf{+3.71 [2.20, 5.22]} & \textbf{+11.39 [6.60, 16.17]} & \textbf{-1.33 [-2.64, -0.02]} & \textbf{+4.22 [3.26, 5.18]} & -1.71 [-4.40, 0.98] \\
        \bottomrule
    \end{tabular}%
    }
    \vspace{-1.8em}
\end{table*}

\subsection{Semantic-Target Sensitivity}
\label{subsec:semantic_target_sensitivity}
We test whether the distinction between geometric stability and semantic recoverability is driven by foreground-volume-tertile targets. We restrict the semantic-transfer comparison to annotation-derived targets in AbdomenAtlas, KiTS23, and TS-MR, excluding the ImageCAS and JHU-Stroke foreground-volume-tertile targets. This is a restriction of the dataset set, not a replacement of the classifiers or labels within a retained dataset.

For CLS, the paired SA--REL advantages are $3.7\;[2.2,5.2]$, $11.4\;[6.6,16.2]$, and $4.2\;[3.3,5.2]$ percentage points, respectively (Table~\ref{tab:acquisition_paired_analysis}). Each adjusted interval excludes zero. REL retains lower drift on AbdomenAtlas and KiTS23, while CKA depends on the dataset. Thus, SA's semantic-transfer advantage persists on these retained datasets even where REL provides lower drift: the stability--recoverability distinction does not depend solely on volume-tertile targets. It remains a property-specific comparison; heterogeneous semantic targets are not pooled into one cross-dataset transfer score. 

\begin{table}[!t]
    \centering
    \scriptsize
    \setlength{\tabcolsep}{5pt}
    \caption{Classifier sensitivity of semantic and anatomical transfer. For each comparison $A$--$B$, entries report $\mathrm{BA}_{A}-\mathrm{BA}_{B}$ in percentage points (pp) over evaluation seeds 42, 123, and 456. Positive effects favor the first checkpoint ($A$); negative effects favor the second ($B$). Brackets give Bonferroni-adjusted 95\% bootstrap confidence intervals (50,000 resamples), with adjustment across REL--ISO, SA--ISO, and SA--REL within each module--dataset--feature--classifier setting. Intervals containing zero establish neither separation nor equivalence.}
    \label{tab:classifier_sensitivity}
    \begin{tabular}{@{}lllrr@{}}
        \toprule
        Dataset & Feature & Comparison & \shortstack{Logistic $C=1$ (pp) [CI]} & \shortstack{Nearest centroid (pp) [CI]} \\
        \midrule
        \multicolumn{5}{l}{\cellcolor{lightgraybg}\textbf{Semantic transfer}} \\
        \addlinespace[2pt]
        AbdomenAtlas & CLS & REL--ISO & $+3.30\,[2.21,4.57]$ & $+5.53\,[4.90,6.28]$ \\
         & & SA--ISO & $+6.82\,[5.41,8.41]$ & $+7.23\,[6.52,8.01]$ \\
         & & SA--REL & $+3.52\,[1.79,5.53]$ & $+1.71\,[1.48,1.96]$ \\
        \addlinespace[2pt]
        AbdomenAtlas & GAP & REL--ISO & $+5.33\,[2.61,7.70]$ & $+9.22\,[8.23,10.19]$ \\
         & & SA--ISO & $+7.32\,[5.15,9.15]$ & $+10.64\,[9.56,11.83]$ \\
         & & SA--REL & $+1.99\,[1.23,3.02]$ & $+1.42\,[1.21,1.69]$ \\
        \addlinespace[2pt]
        KiTS23 & CLS & REL--ISO & $-8.52\,[-13.19,-2.82]$ & $-6.35\,[-8.32,-4.15]$ \\
         & & SA--ISO & $+3.98\,[1.06,6.82]$ & $-3.43\,[-9.62,2.59]$ \\
         & & SA--REL & $+12.50\,[6.90,16.84]$ & $+2.92\,[-2.00,7.84]$ \\
        \addlinespace[2pt]
        KiTS23 & GAP & REL--ISO & $+7.66\,[1.87,14.45]$ & $+0.68\,[-4.38,5.00]$ \\
         & & SA--ISO & $+5.02\,[1.33,8.61]$ & $+1.02\,[-4.79,6.65]$ \\
         & & SA--REL & $-2.64\,[-7.33,0.32]$ & $+0.33\,[-1.83,2.59]$ \\
        \addlinespace[2pt]
        TS-MR & CLS & REL--ISO & $+12.68\,[11.66,13.49]$ & $+10.93\,[9.78,12.14]$ \\
         & & SA--ISO & $+17.21\,[15.99,18.12]$ & $+15.38\,[14.32,16.15]$ \\
         & & SA--REL & $+4.53\,[3.85,5.09]$ & $+4.44\,[3.49,5.36]$ \\
        \addlinespace[2pt]
        TS-MR & GAP & REL--ISO & $+16.06\,[14.86,17.46]$ & $+15.80\,[14.02,18.13]$ \\
         & & SA--ISO & $+19.17\,[17.86,20.44]$ & $+20.47\,[19.10,21.97]$ \\
         & & SA--REL & $+3.11\,[2.09,4.43]$ & $+4.67\,[3.62,5.46]$ \\
        \midrule
        \multicolumn{5}{l}{\cellcolor{lightgraybg}\textbf{Anatomical transfer}} \\
        \addlinespace[2pt]
        TotalSeg CT--MR & CLS & REL--ISO & $+7.82\,[1.25,13.69]$ & $+9.80\,[3.77,15.56]$ \\
         & & SA--ISO & $+9.18\,[1.24,16.54]$ & $+11.09\,[2.40,19.35]$ \\
         & & SA--REL & $+1.36\,[1.76,7.63]$ & $+1.29\,[0.13,7.95]$ \\
        \addlinespace[2pt]
        TotalSeg CT--MR & GAP & REL--ISO & $+29.20\,[20.29,38.83]$ & $+32.54\,[25.30,39.50]$ \\
         & & SA--ISO & $+33.99\,[24.26,44.21]$ & $+35.85\,[26.74,45.09]$ \\
         & & SA--REL & $+4.79\,[0.31,9.31]$ & $+3.31\,[2.37,7.69]$ \\
        \addlinespace[2pt]
        MMWHS & CLS & REL--ISO & $+9.66\,[-0.79,21.03]$ & $+12.04\,[2.91,22.35]$ \\
         & & SA--ISO & $+11.77\,[3.31,22.62]$ & $+13.10\,[4.89,23.02]$ \\
         & & SA--REL & $+2.12\,[-10.71,15.21]$ & $+1.06\,[-8.07,12.70]$ \\
        \addlinespace[2pt]
        MMWHS & GAP & REL--ISO & $+36.51\,[28.57,46.16]$ & $+41.01\,[27.65,53.17]$ \\
         & & SA--ISO & $+39.15\,[28.97,50.40]$ & $+44.97\,[29.23,60.85]$ \\
         & & SA--REL & $+2.65\,[1.97,10.32]$ & $+3.97\,[2.69,12.30]$ \\
        \addlinespace[2pt]
        AMOS22 & CLS & REL--ISO & $+12.34\,[3.31,21.83]$ & $+12.60\,[3.51,22.21]$ \\
         & & SA--ISO & $+16.35\,[6.36,26.05]$ & $+16.93\,[7.01,26.63]$ \\
         & & SA--REL & $+4.02\,[2.10,10.30]$ & $+4.33\,[3.23,9.87]$ \\
        \addlinespace[2pt]
        AMOS22 & GAP & REL--ISO & $+37.69\,[27.55,49.61]$ & $+41.63\,[31.47,51.52]$ \\
         & & SA--ISO & $+44.22\,[34.33,55.81]$ & $+50.05\,[39.99,60.30]$ \\
         & & SA--REL & $+6.53\,[2.25,11.66]$ & $+8.42\,[1.98,13.74]$ \\
        \bottomrule
    \end{tabular}
    \vspace{-2.0em}
\end{table}

\subsection{Classifier Sensitivity}
\label{subsec:classifier_sensitivity}
We examine whether semantic-transfer and anatomical transfer comparisons depend on classifier choice (Table~\ref{tab:classifier_sensitivity}). Within this sensitivity analysis, features, cases, targets, partitions, and training-only standardization remain fixed across classifier settings for each checkpoint. We vary logistic regression's inverse regularization strength, $C\in\{0.1,1,10\}$, and compare it with nearest-centroid classification, which requires no iterative optimization. For each binary semantic target, positive and negative centroids are computed from standardized source-training features; target-spacing test features are classified by Euclidean distance to these centroids. The table reports all three pairwise contrasts for $C=1$ and nearest centroid. The $C=1$ baselines are recomputed using the benchmark protocol and verified against its evaluation routines, rather than taken from the original estimates retained in the main tables.

Regularization changes preserve the mean ISO/REL/SA ordering across all six dataset--feature combinations in each module. Nearest-centroid classification preserves that ordering in four of six semantic-transfer combinations and all six anatomical-transfer combinations. The semantic ordering changes only on KiTS23; AbdomenAtlas and TS-MR retain $\mathrm{SA}>\mathrm{REL}>\mathrm{ISO}$, with positive adjusted intervals for all three contrasts under both classifiers and feature types. In anatomical transfer, adjusted intervals favor SA over ISO throughout and REL over ISO except for MMWHS--CLS with logistic regression. Across the three organ transfer datasets, two feature types, and two classifiers, SA also achieves higher mean accuracy than REL in all twelve settings, with positive adjusted intervals in ten. The two unresolved SA--REL comparisons are MMWHS--CLS with logistic regression and nearest centroid. Thus, the mean accuracy ordering is consistent across classifiers, while statistical separation remains comparison-specific.

\subsection{Support-Budget Sensitivity}
\label{subsec:support_budget_sensitivity}
To test whether the organ transfer comparison depends on the two-sample support budget, we evaluate CLS probe BA at one, two, four, and eight support samples per organ and modality, using case-disjoint sampling shared across checkpoints within each budget. Table~\ref{tab:support_budget_sensitivity} averages dataset scores with equal weight. This additional sweep evaluates support-budget dependence separately
from the estimates retained in the main table. We stop at eight shots, which retains 12 MMWHS query cases per organ and modality; larger budgets further reduce the query pool. Query cases change as support increases, so comparisons are controlled within each budget rather than isolating the gain from additional support across rows.
\begin{table}[!t]
    \centering
    \scriptsize
    \caption{CLS organ transfer support-budget sweep. Probe BA is averaged over seeds (42,123,456) and then equally over TotalSegmentator CT--MR, MMWHS, and AMOS22; values are \(\times\)100. The third column counts datasets on which SA has the highest mean. Query counts are per organ and modality.}
    \label{tab:support_budget_sensitivity}
    \begin{tabular}{@{}rrrr@{}}
        \toprule
        Support & 3DINO/REL/SA probe BA & SA highest mean
        & Query cases: TS/MMWHS/AMOS22 \\
        \midrule
        1 & 54.8/55.3/56.4 & 1/3 & 85/19/54 \\
        2 & 64.7/66.1/68.7 & 3/3 & 84/18/53 \\
        4 & 71.2/75.1/77.5 & 3/3 & 82/16/51 \\
        8 & 76.9/80.0/81.5 & 3/3 & 77--78/12/47 \\
        \bottomrule
\end{tabular}
    \vspace{-1.5em}
\end{table}

The mean ordering is SA $>$ REL $>$ 3DINO throughout. At one shot, SA has the highest mean on one of the three datasets; at two, four, and eight shots, it has the highest mean on every dataset. SA's aggregate mean advantage is thus not confined to the primary two-sample budget, while its dataset-level lead becomes consistent only at the tested budgets of two or more samples per organ and modality.

\subsection{Protocol Stability}
\label{subsec:protocol_stability}
We test whether policy rankings depend on how the evaluation inputs and anatomical categories are defined. We vary acquisition preprocessing, organ-region definition, and organ-label granularity while holding the evaluated checkpoint and the eligible cohort fixed within each analysis. The controlled comparisons use ISO, REL, and SA; the main ranking summary uses CLS. Table~\ref{tab:protocol_stability} compares each altered protocol with its specified reference. Kendall's $\tau$ measures agreement of the complete three-checkpoint ordering ($1$ means identical ordering); leader retention counts settings in which the reference winner remains first. These quantities describe ranking stability, not invariance of absolute scores or statistical equivalence of checkpoints.

\paragraph{Acquisition and preprocessing.}
To assess dependence on the resampling pipeline and physical coverage, we vary preprocessing while retaining the same scans. AbdomenAtlas, KiTS23, and TS-MR are evaluated for drift, CKA, and semantic transfer at the moderate and severe target conditions. We separately disable anti-aliasing, replace trilinear interpolation with nearest-neighbor interpolation, or use a centered physical FOV of $192\times192\times192$ mm. Other protocol settings remain fixed. The 54 comparisons are three datasets $\times$ three metrics $\times$ two target conditions $\times$ three evaluation seeds. Moderate and severe refer to each dataset's predefined target spacings, not common absolute scanner settings.

\paragraph{Region definition.}
The margin and masking interventions distinguish dependence on crop extent from dependence on surrounding image content; erosion further tests sensitivity to image content near the organ boundary. AMOS22 and MMWHS use five predefined $96^3$ crop surfaces: the original bounding box with a five-voxel margin and retained context (M5), a zero-margin box with retained context (M0), the same two boxes with outside-organ content zeroed, and an organ-only M5 crop with a 1-mm erosion. Margins alter crop extent and surrounding context, not the original organ mask. Erosion is defined by $\operatorname{EDT}(M_{\mathrm{padded}};s)>1\,\mathrm{mm}$, with the physical voxel spacing $s$, and is applied before axis permutation. The bounding box always comes from the original mask; erosion changes only the image content retained inside it. Original masks contain at least 100 voxels and transformed masks must remain nonempty, preserving the manifest cohort.

Sampling plans store exact bidirectional query/target IDs and case-disjoint support/query IDs and are shared across the three checkpoints and crop surfaces for each seed. Table~\ref{tab:region_definition_sensitivity} reports the crop-margin and context comparisons.
\begin{table*}[!t]
    \centering
    \scriptsize
    \setlength{\tabcolsep}{4pt}
    \caption{Crop-margin and context sensitivity on AMOS22 and MMWHS. The reported CLS analysis uses seeds 42, 123, and 456; scores are equal-weight dataset means $\times100$, in ISO/REL/SA order. Pairwise entries are SA--REL differences in percentage points with dataset-stratified, Bonferroni-adjusted 95\% intervals across the three checkpoint comparisons. M5/M0 denote five-/zero-voxel margins.}
    \label{tab:region_definition_sensitivity}
    \resizebox{\textwidth}{!}{%
    \begin{tabular}{@{}lrrrr@{}}
        \toprule
        Surface & mAP: ISO/REL/SA & SA--REL & Top-1: ISO/REL/SA & SA--REL \\
        \midrule
        M5 + context & 15.3/33.2/42.7 & $+9.5\;[8.6,10.4]$ & 14.8/44.5/60.5 & $+16.0\;[13.3,18.7]$ \\
        M0 + context & 15.5/31.5/40.4 & $+8.8\;[7.8,9.8]$ & 12.9/38.7/52.1 & $+13.4\;[10.8,16.1]$ \\
        M5 + organ-only & 16.7/24.0/26.4 & $+2.4\;[1.6,5.4]$ & 18.6/27.9/31.8 & $+3.9\;[1.7,6.1]$ \\
        M0 + organ-only & 17.8/22.2/23.7 & $+1.5\;[0.6,2.6]$ & 18.9/31.3/32.5 & $+1.2\;[1.0,2.5]$ \\
        \bottomrule
    \end{tabular}%
    }
    \vspace{-1.5em}
\end{table*}

All four margin/context settings retain the mean ordering $\mathrm{SA}>\mathrm{REL}>\mathrm{ISO}$ for mAP and Top-1 (Table~\ref{tab:region_definition_sensitivity}). Masking context lowers the REL and SA scores and narrows their differences, but all adjusted SA--REL intervals remain positive. Thus, SA's advantage persists within the organ-only crop construction, while its magnitude depends on surrounding context; though stable ordering does not imply context-independent representations. The 1-mm erosion retains five of six reference leaders with mean Kendall's $\tau=0.89$; only the MMWHS probe BA leader changes from REL to ISO (Table~\ref{tab:protocol_stability}). This supports stability under the tested boundary change, rather than invariance to mask definition.

\paragraph{Organ granularity.}
The \emph{Fine-organ $\rightarrow$ organ-family labels} row in Table~\ref{tab:protocol_stability} tests whether the ISO/REL/SA ranking persists when left/right distinctions are removed from the evaluation labels. We reuse the baseline features without additional image processing or encoder inference. In AMOS22, left/right kidneys and adrenal glands are merged into kidney and adrenal-gland classes, reducing 13 classes to 11. In MMWHS, left/right atria and ventricles are merged into atrium and ventricle, reducing seven classes to five. Other labels and source-case identifiers remain unchanged. At most one crop per source case, class, and modality is retained before balanced sampling, preventing merged laterality labels from duplicating a patient's contribution. This tests a defined laterality collapse, not arbitrary organ-system groupings.

Raw fine-to-family score increases are not interpreted as improved representations because fewer classes change retrieval and classification difficulty. We therefore compare checkpoint rankings within each label definition rather than treating higher coarse-label scores as gains from pretraining.

This row aggregates mAP, Top-1, and probe balanced-accuracy rankings on AMOS22 and MMWHS after averaging metric scores over evaluation seeds. The reference leader is retained in five of these six dataset--metric comparisons, with mean Kendall's $\tau=0.78$. Most leading-policy conclusions therefore persist after removing laterality distinctions, but the remaining leader change and imperfect ordering agreement show that label granularity can affect the comparative conclusion.
\begin{table}[t]
    \centering
    \scriptsize
    \caption{CLS ISO/REL/SA ranking stability over seeds 42, 123, and 456. Acquisition rows average 54 seed-specific dataset--metric--severity comparisons. Organ granularity and erosion use six dataset--metric comparisons after averaging seeds: AMOS22/MMWHS $\times$ mAP/Top-1/probe BA. Each row uses its own reference; erosion is compared with unmodified organ-only M5 masks. Retention percentages are rounded from the displayed counts. M5 denote 5-voxel margins.}
    \label{tab:protocol_stability}
    \begin{tabular}{@{}lrr@{}}
        \toprule
        Controlled change & Mean Kendall $\tau$ & Reference leader retained \\
        \midrule
        Disable anti-aliasing & 0.82 & 51/54 (94\%) \\
        Trilinear $\rightarrow$ nearest & 0.73 & 48/54 (89\%) \\
        Centered fixed physical FOV & 0.70 & 46/54 (85\%) \\
        Fine-organ $\rightarrow$ organ-family labels & 0.78 & 5/6 (83\%) \\
        Organ-only M5 $\rightarrow$ 1-mm mask erosion & 0.89 & 5/6 (83\%) \\
        \bottomrule
    \end{tabular}
    \vspace{-1.5em}
\end{table}

Together, the ablations support stability of most leading-policy conclusions under the tested changes, while the crop/context results show that stable rankings can coexist with changes in scores and effect sizes. Crop/context, interpolation, FOV, and label definitions therefore remain part of the benchmark specification rather than interchangeable implementation choices. Controlled resampling does not simulate scanner physics or the full variability of clinical acquisition.

\subsection{Organ, Modality-Direction, and Spacing-Severity Analysis}
\label{subsec:organ_modality_severity}
We analyze CLS SA--REL effects from the additional frozen-feature evaluations separately from the estimates retained in the main tables. Table~\ref{tab:organ_modality_severity} resolves spacing severity and retrieval direction; severity-specific semantic transfer compares reference spacing with each moderate or severe target, rather than averaging all six transfer directions. Averaged equally across the five acquisition datasets, REL has lower mean drift and higher mean CKA at both target severities. Semantic transfer favors SA on AbdomenAtlas, KiTS23, and TS-MR at both severities, REL on ImageCAS, and changes direction on JHU-Stroke. Thus, the semantic-transfer advantage remains dataset-specific at each controlled spacing severity.

For retrieval, SA has higher mean mAP in both AMOS22 directions and in MMWHS CT$\rightarrow$MR; REL is higher in MMWHS MR$\rightarrow$CT. In the latter direction, SA leads on four of seven organs, but REL's larger gains on the other organs determine the aggregate. Organ counts therefore complement, rather than replace, the magnitude of the mean difference. Table~\ref{tab:per_organ_retrieval_effects} reports every organ-level difference.
\begin{table*}[t]
    \centering
    \scriptsize
    \setlength{\tabcolsep}{4pt}
    \caption{CLS SA--REL effects over seeds 42, 123, and 456. A named checkpoint followed by $+x$ denotes its advantage in the metric's preferred direction: for example, REL $+6.6$ is 6.6 points lower drift. Scores are scaled by 100. Acquisition drift and CKA are equal-weight five-dataset means; semantic transfer remains dataset-specific. Moderate/severe are controlled resampling targets, not measured scanner categories. Retrieval counts denote organs with higher SA mean mAP and are descriptive, not significance counts.}
    \label{tab:organ_modality_severity}
    \resizebox{\textwidth}{!}{%
    \begin{tabular}{@{}lrrp{0.64\textwidth}@{}}
        \toprule
        \multicolumn{4}{l}{\textbf{Acquisition: controlled spacing severity}} \\
        Condition & Drift $\downarrow$ & CKA $\uparrow$ & Transfer: signed SA--REL effect \\
        \midrule
        Moderate & REL $+6.6$ & REL $+1.4$ & AbdomenAtlas $+2.8$; KiTS23 $+4.8$; TS-MR $+4.0$; ImageCAS $-2.8$; JHU $-3.9$ \\
        Severe & REL $+7.5$ & REL $+0.7$ & AbdomenAtlas $+1.1$; KiTS23 $+11.6$; TS-MR $+4.7$; ImageCAS $-3.0$; JHU $+1.1$ \\
        \midrule
        \multicolumn{4}{l}{\textbf{Retrieval: modality direction and organ-level counts}} \\
        Dataset & \multicolumn{2}{c}{CT$\rightarrow$MR mAP $\uparrow$} & MR$\rightarrow$CT mAP $\uparrow$ \\
        \midrule
        AMOS22 & \multicolumn{2}{c}{SA $+4.9$; SA higher on 8/13 organs} & SA $+4.3$; SA higher on 10/13 organs \\
        MMWHS & \multicolumn{2}{c}{SA $+3.9$; SA higher on 5/7 organs} & REL $+1.4$; SA higher on 4/7 organs \\
        \bottomrule
    \end{tabular}%
    }
    \vspace{-0.7em}
\end{table*}

The organ-level results clarify the MMWHS direction reversal: the ascending aorta and right atrium favor SA in CT$\rightarrow$MR but show larger REL-favoring differences in MR$\rightarrow$CT. These differences outweigh the positive SA effects on four other organs in that direction. Together with the dataset-specific severity patterns, this shows why neither a bidirectional mean nor a count of favorable organs fully describes the policy comparison. These descriptive breakdowns identify where SA helps and where it does not; they do not establish that spacing sensitivity causes cross-modal performance.
\begin{table*}[t]
    \centering
    \scriptsize
    \caption{Complete CLS organ-level SA--REL mAP differences, averaged over seeds 42, 123, and 456 and expressed in percentage points. Positive values favor SA; negative values favor REL. These are descriptive mean differences without per-organ significance claims. The equal-organ means reproduce the directional effects in Table~\ref{tab:organ_modality_severity}.}
    \label{tab:per_organ_retrieval_effects}
    \begin{tabular}{@{}lrr|lrr@{}}
        \toprule
        \multicolumn{3}{c|}{\textbf{AMOS22}} &
        \multicolumn{3}{c}{\textbf{MMWHS}} \\
        \midrule
        Organ & CT$\rightarrow$MR & MR$\rightarrow$CT &
        Organ & CT$\rightarrow$MR & MR$\rightarrow$CT \\
        \midrule
        Adrenal gland left & $-4.96$ & $+9.65$ &
        Ascending aorta & $+6.23$ & $-25.40$ \\
        Adrenal gland right & $-18.84$ & $+5.49$ &
        Left atrium & $+15.77$ & $+12.27$ \\
        Aorta & $+21.37$ & $+11.41$ &
        Left ventricle & $+1.45$ & $+10.15$ \\
        Duodenum & $-2.60$ & $-0.11$ &
        Myocardium & $-1.31$ & $+6.18$ \\
        Esophagus & $+14.99$ & $+15.53$ &
        Pulmonary artery & $-5.04$ & $+4.72$ \\
        Gallbladder & $-9.30$ & $+2.43$ &
        Right atrium & $+0.95$ & $-16.86$ \\
        Inferior vena cava & $+29.50$ & $+7.36$ &
        Right ventricle & $+9.53$ & $-0.56$ \\
        Kidney left & $+13.81$ & $+2.74$ & & & \\
        Kidney right & $+8.61$ & $+3.94$ & & & \\
        Liver & $+4.82$ & $+1.51$ & & & \\
        Pancreas & $+4.60$ & $+4.22$ & & & \\
        Spleen & $+5.46$ & $-5.36$ & & & \\
        Stomach & $-3.24$ & $-2.47$ & & & \\
        \bottomrule
    \end{tabular}
    \vspace{-1.7em}
\end{table*}

\subsection{Pretraining-Unseen Evaluation}
\label{subsec:pretraining_unseen_evaluation}
We examine whether the policy comparisons extend to datasets absent from Med3DINO pretraining. The exposure-restricted comparison includes KiTS23 and TS-MR for Acquisition Robustness, and MMWHS and AMOS22 for cross-modal and anatomical transfer. TotalSegmentator CT--MR is excluded because its CT component is pretraining-seen; HECKTOR is also pretraining-seen and is not included. CHAOS is kept separate because its identifier-based liver pairing between different CT and MR patients is not comparable to organ-category retrieval. No corresponding unseen-pretraining claim is made for external encoders whose training-data exposure differs.

Table~\ref{tab:pretraining_unseen_summary} summarizes the reported restricted analysis. Ranks order only ISO/REL/SA within each dataset and metric and then average equally across the two datasets. Paired wins require a Bonferroni-adjusted interval excluding zero. The table separates mean ordering from paired evidence and does not combine metrics into a global ranking. Its paired outcomes belong to this restricted analysis, rather than being substituted for intervals from the primary or classifier-sensitivity evaluations.
\begin{table}[t]
    \centering
    \scriptsize
    \caption{Reported CLS pretraining-unseen summary over seeds 42, 123, and 456. Rank 1 is best; mean ranks give equal weight to the two eligible datasets for each module. The last column counts supported SA--REL outcomes out of two datasets. NS (Not Supported) denotes an adjusted interval containing zero, not equivalence.}
    \label{tab:pretraining_unseen_summary}
    \begin{tabular}{@{}lrr@{}}
        \toprule
        Metric & ISO/REL/SA mean rank $\downarrow$ & SA--REL outcomes \\
        \midrule
        Acquisition drift & 2.5/1.0/2.5 & REL 1; NS 1 \\
        Acquisition CKA & 1.5/2.5/2.0 & REL 1; SA 1 \\
        Acquisition transfer & 2.5/2.5/1.0 & SA 2 \\
        Cross-modal mAP & 3.0/2.0/1.0 & SA 2 \\
        Cross-modal Top-1 & 3.0/2.0/1.0 & SA 2 \\
        Linear probe BA & 3.0/2.0/1.0 & SA 2 \\
        \bottomrule
    \end{tabular}
    \vspace{-1.8em}
\end{table}

Within this restricted analysis, SA ranks first for cross-modal mAP, Top-1, and probe BA, with adjusted SA--REL intervals favoring SA on both eligible datasets for each metric (Table~\ref{tab:pretraining_unseen_summary}). Acquisition outcomes remain metric-specific: REL has the best average drift rank, with a supported advantage on one dataset and no supported separation on the other; CKA favors REL on KiTS23 and SA on TS-MR, while semantic transfer
favors SA on both datasets. SA's advantages in retrieval and probe BA therefore extend to these pretraining-unseen datasets, rather than depending exclusively on pretraining-seen collections.

\subsection{CT--PET Retrieval Extension}
\label{subsec:ct_pet_extension}
CT--MR provides the primary shared multi-organ evaluation, but the retrieval protocol only requires a defined target vocabulary supported in both modalities. To test its applicability to another modality pair and label vocabulary, we evaluate HECKTOR-2025~\cite{hecktor2025} CT--PET with two categories: primary tumor and metastatic lymph nodes. Masks are mapped into each modality's physical image grid with nearest-neighbor label interpolation. Query and target pools are balanced by category, and every target from the query patient's case is excluded. Retrieval therefore cannot succeed by directly matching a patient's registered CT/PET pair. Encoders remain frozen, and CT$\rightarrow$PET and PET$\rightarrow$CT scores are averaged equally. 

From 678 complete CT/PET pairs, we select 500 category-balanced crops representing 229 patients. Each category contributes 125 crops per modality; each retrieval direction samples 100 queries and 100 targets per category, totaling 200 queries and 200 targets. After same-patient exclusion, queries retain 198--200 candidate targets.

Table~\ref{tab:ct_pet_extension} reports higher mean mAP and Top-1 for SA than REL and ISO with both feature types. The agreement across CLS and GAP supports a consistent descriptive SA advantage in this cross-patient lesion-category task. For these patient-excluded pools, random ranking has expected mAP/Top-1 of 51.07/49.83 on the reported scale. Scores are interpreted relative to this reference and are not pooled with multi-organ CT--MR results. The HECKTOR resource is pretraining-seen for Med3DINO. These results extend the retrieval evaluation beyond CT--MR and organ labels, without establishing general modality invariance. % or equivalence to the multi-organ task.
\begin{table}[!t]
    \centering
    \scriptsize
    \setlength{\tabcolsep}{4pt}
    \caption{HECKTOR-2025 balanced lesion-category CT$\leftrightarrow$PET retrieval. Entries are mAP/Top-1, expressed as mean$\pm$sample SD over seeds 42, 123, and 456 and scaled by 100.}
    \label{tab:ct_pet_extension}
    \begin{tabular}{@{}lrrr@{}}
        \toprule
        Feature & ISO (mAP/Top-1) & REL (mAP/Top-1) & SA (mAP/Top-1) \\
        \midrule
        CLS & $51.53\pm0.09$ / $47.83\pm1.04$ & $52.21\pm0.11$ / $49.83\pm0.38$ & $53.73\pm0.24$ / $54.17\pm2.67$ \\
        GAP & $51.97\pm0.09$ / $48.92\pm1.53$ & $52.78\pm0.06$ / $53.00\pm0.90$ & $53.69\pm0.10$ / $55.08\pm3.17$ \\
        \bottomrule
    \end{tabular}
    \vspace{-1.5em}
\end{table}

\section{Implementation and Reproducibility}
\label{sec:implementation_reproducibility_benchmarking}
REFORM-3D is provided as a standalone Python package, with the benchmark implementation under \highlightLR{\texttt{src/reform3d/}}.

\subsection{Experiment Identification and Resources}
\label{subsec:code_structure}
Each evaluation is specified by its dataset and sampled cohort, checkpoint, frozen-feature interface, preprocessing, task settings, and evaluation seed. Dataset construction and metric definitions are given in Sections~\ref{sec:benchmark_dataset_construction} and~\ref{sec:benchmark_protocol_metric}; additional evaluation settings are specified in Section~\ref{sec:additional_results_and_extended_tables}. Reproducing a reported comparison requires its corresponding manifests, checkpoint weights, and configuration.

Table~\ref{tab:reproducibility_resources} summarizes the repository's implementation and reproducibility resources. YAML configurations specify evaluation inputs; \highlightLR{\texttt{reform3d check}} validates them, and \highlightLR{\texttt{reform3d run}} executes the configured modules. External encoders retain their declared preprocessing and frozen-feature interfaces. Installation instructions, configuration examples, and encoder-specific requirements are documented in the repository.
\begin{table*}[ht]
    \centering
    \scriptsize
    \setlength{\tabcolsep}{4pt}
    \caption{Implementation and reproducibility resources in the REFORM-3D repository. Paths are relative to the repository root.}
    \label{tab:reproducibility_resources}
    \begin{tabular}{@{}lp{0.59\textwidth}p{0.22\textwidth}@{}}
        \toprule
        Resource & Repository artifact & Role \\
        \midrule
        \rowcolor{yellow!12}
        Data and cohorts &
        \url{docs/data.md}; \url{src/reform3d/prepare.py} &
        Dataset inputs, manifest preparation, and cohort validation. \\
        \rowcolor{blue!8}
        Evaluation &
        \url{src/reform3d/cli.py}; \url{configs/} &
        Configured input validation and module execution. \\
        \rowcolor{green!8}
        Encoders &
        \url{reproduction/checkpoints.json}; \url{docs/encoders.md} &
        Checkpoint identities, available weight hashes, interfaces, and dependencies. \\
        \rowcolor{purple!8}
        Environment &
        \url{requirements.txt}; \url{pyproject.toml} &
        Tested dependency versions and package installation. \\
        \rowcolor{cyan!10}
        Usage &
        \url{docs/usage.md} &
        Configuration, cache reuse, outputs, and failure handling. \\
        \bottomrule
    \end{tabular}
    \vspace{-1.7em}
\end{table*}

\subsection{Manifests, Caches, and Outputs}
\label{subsec:manifests_caches_outputs}
Manifests record eligible source images, anatomical targets, modalities, and source-case identifiers. The YAML \texttt{seeds} field selects repeated evaluations, with results stored under \highlightLR{\path{runs/<run_name>/seeds/seed_<N>/}}. Shared deterministic crop caches avoid repeated image preprocessing. Checkpoint- and feature-specific bundles remain isolated, and sensitivity bundles additionally record manifest and preprocessing signatures. Before reusing completed features or metrics, the standalone runner verifies configuration compatibility and their recorded SHA-256 checksums.

Region-definition and organ-granularity experiments save reusable sampling plans, verify unique sample IDs and exact cohort coverage, and reject incomplete bundles before aggregation. Their analyses retain query/organ-level records for reconstructing headline metrics and paired effects.

% \subsection{Seeds and Uncertainty}
% \label{subsec:seeds_determinism}
% Evaluation seeds 42, 123, and 456 control sampling or classifier partitions, not independent pretraining runs. Bootstrap sampling and multiplicity are specified in Section~\ref{subsec:uncertainty_protocol}; standard deviations, paired intervals, and checkpoint-ranking agreement measure different quantities and are not interchangeable. Region-definition paired analyses use 50,000 bootstrap resamples with seed 2026. Bonferroni adjustment is applied separately to the three checkpoint comparisons and four predefined region comparisons: M0 context versus M5 context, organ-only M5 versus M5 context, organ-only M0 versus M5 context, and eroded M5 versus organ-only M5.

\subsection{Data Access and Release Documentation}
\label{subsec:data_access_release}
The source inventory in Section~\ref{subsec:dataset_breakdown} describes corpus composition; it does not authorize redistribution of provider-controlled scans or annotations. Users must obtain restricted sources from their original providers under the applicable terms. Reconstruction scripts operate on locally obtained data, and any redistribution of derived assets remains subject to those source terms.

Release documentation will specify dataset versions, cohort-selection rules, organ mappings, evaluation seeds, checkpoint sources, applicable terms, and available SHA-256 hashes. Croissant metadata and a responsible-AI (RAI) benchmark card will document dataset composition, pretraining exposure, intended use, access constraints, and limitations.

\subsection{Compute Accounting}
\label{subsec:compute_resources}
The controlled Med3DINO comparison uses the common 125K-update pretraining budget. Evaluation separates image preprocessing, frozen-feature extraction, and CPU classifier/statistical analysis. Features are reused across evaluation seeds when their cohort and preprocessing are unchanged; organ-granularity analysis reuses baseline features without additional encoder inference. Pretraining was conducted on a 16-node cluster with four NVIDIA A100 GPUs per node, with a training duration of eight days per checkpoint. Evaluation used two NVIDIA A100 GPUs and typically required 1--2 days per dataset, depending on dataset size and evaluation protocol.

\subsection{Downstream Segmentation Protocol}
\label{subsec:downstream_segmentation_protocol}
For downstream segmentation, we keep the pretrained encoder frozen and train a ViTAdapterUNETR decoder, following 3DINO~\cite{3dinov2} with each dataset's labeled training set. Training uses AdamW with an initial learning rate of $10^{-4}$ and a warmup-cosine schedule for 100 epochs, with early stopping after 30 validation evaluations without improvement. Crop dimensions, batch size, and iterations per epoch follow the dataset-specific run settings. Most datasets use combined Dice and cross-entropy loss; BraTS-GLI and ISLES22 use sigmoid Dice loss. We select the checkpoint with the highest validation Dice and evaluate it on the test split using sliding-window inference. Dice is averaged over test scans, while HD95 aggregation excludes non-finite per-scan values.

\newpage
\input{checklist.tex}

\end{document}

%% file: checklist.tex
\section*{NeurIPS Paper Checklist}

%%% BEGIN INSTRUCTIONS %%%
The checklist is designed to encourage best practices for responsible machine learning research, addressing issues of reproducibility, transparency, research ethics, and societal impact. Do not remove the checklist: {\bf The papers not including the checklist will be desk rejected.} The checklist should follow the references and follow the (optional) supplemental material.  The checklist does NOT count towards the page
limit. 

Please read the checklist guidelines carefully for information on how to answer these questions. For each question in the checklist:
\begin{itemize}
    \item You should answer \answerYes{}, \answerNo{}, or \answerNA{}.
    \item \answerNA{} means either that the question is Not Applicable for that particular paper or the relevant information is Not Available.
    \item Please provide a short (1--2 sentence) justification right after your answer (even for \answerNA). 
   % \item {\bf The papers not including the checklist will be desk rejected.}
\end{itemize}

{\bf The checklist answers are an integral part of your paper submission.} They are visible to the reviewers, area chairs, senior area chairs, and ethics reviewers. You will also be asked to include it (after eventual revisions) with the final version of your paper, and its final version will be published with the paper.

The reviewers of your paper will be asked to use the checklist as one of the factors in their evaluation. While \answerYes{} is generally preferable to \answerNo{}, it is perfectly acceptable to answer \answerNo{} provided a proper justification is given (e.g., error bars are not reported because it would be too computationally expensive'' or ``we were unable to find the license for the dataset we used''). In general, answering \answerNo{} or \answerNA{} is not grounds for rejection. While the questions are phrased in a binary way, we acknowledge that the true answer is often more nuanced, so please just use your best judgment and write a justification to elaborate. All supporting evidence can appear either in the main paper or the supplemental material, provided in appendix. If you answer \answerYes{} to a question, in the justification please point to the section(s) where related material for the question can be found.

IMPORTANT, please:
\begin{itemize}
    \item {\bf Delete this instruction block, but keep the section heading ``NeurIPS Paper Checklist"},
    \item  {\bf Keep the checklist subsection headings, questions/answers and guidelines below.}
    \item {\bf Do not modify the questions and only use the provided macros for your answers}.
\end{itemize}

%%% END INSTRUCTIONS %%%

\begin{enumerate}

\item {\bf Claims}
    \item[] Question: Do the main claims made in the abstract and introduction accurately reflect the paper's contributions and scope?
    \item[] Answer: \answerYes{} % Replace by \answerYes{}, \answerNo{}, or \answerNA{}.
    % \item[] Justification: The introduction clearly states the paper’s core contributions, proposing REFORM-3D as a representation-centric evaluation framework, defining its three modules (Acquisition Robustness, Cross-Modal Anatomical Alignment, and Anatomical Generalization), and evaluating Med3DINO encoders under different geometric pretraining regimes. These claims directly match the methods, experimental design, and results presented in the paper.
    \item[] Justification: The abstract and introduction state the paper's core contributions: REFORM-3D as a representation-centric framework comprising Acquisition Robustness, Cross-Modal Anatomical Alignment, and Anatomical Transfer; Med3DINO encoders under controlled geometric pretraining regimes; and benchmarking across encoder families. Sections~\ref{sec:benchmark_design} and~\ref{sec:benchmark_results} document the corresponding methods and results. Geometric-policy attribution is restricted to the controlled Med3DINO comparison, while external encoders provide contextual comparisons.

\item {\bf Limitations}
    \item[] Question: Does the paper discuss the limitations of the work performed by the authors?
    \item[] Answer: \answerYes{} % Replace by \answerYes{}, \answerNo{}, or \answerNA{}.
    % \item[] Justification: The paper includes a dedicated Limitations section (Sec.~\ref{sec:limitations}) that discusses the scope of REFORM-3D and clarifies that the framework evaluates specific representation properties under controlled acquisition, modality, and anatomical settings rather than establishing full invariance or replacing downstream evaluation. It also highlights assumptions in the Acquisition Robustness module, noting that controlled spacing perturbations act as geometric stress tests and may not fully capture real-world acquisition variability.
    \item[] Justification: Section~\ref{sec:conclusion} discusses the evaluated representation settings and clarifies that REFORM-3D neither establishes broader invariance nor replaces downstream evaluation. It identifies controlled resampling as geometric stress testing rather than full clinical acquisition realism, and limits the REL--SA findings to the studied architecture and self-supervised objective.

\item {\bf Theory assumptions and proofs}
    \item[] Question: For each theoretical result, does the paper provide the full set of assumptions and a complete (and correct) proof?
    \item[] Answer: \answerNA{} % Replace by \answerYes{}, \answerNo{}, or \answerNA{}.
    \item[] Justification: The paper does not include theoretical results such as theorems or formal proofs. It focuses on defining an evaluation framework, associated metrics, and empirical benchmarking protocols for analyzing frozen 3D representations.

    \item {\bf Experimental result reproducibility}
    \item[] Question: Does the paper fully disclose all the information needed to reproduce the main experimental results of the paper to the extent that it affects the main claims and/or conclusions of the paper (regardless of whether the code and data are provided or not)?
    \item[] Answer: \answerYes{} % Replace by \answerYes{}, \answerNo{}, or \answerNA{}.
    % \item[] Justification: The paper provides detailed descriptions of the REFORM-3D evaluation framework, including dataset construction, evaluation protocols, and metric definitions for all three modules. The supplementary material further includes implementation details, reproducibility guidelines, and dataset composition. Together, these enable reproduction of the main experimental results, and code and pretrained models will be released to facilitate verification.The paper provides detailed descriptions of the REFORM-3D evaluation framework, including dataset construction, evaluation protocols, and metric definitions for all three modules. The supplementary material further includes implementation details, reproducibility guidelines, and dataset composition. Together, these enable reproduction of the main experimental results, and code and pretrained models will be released to facilitate verification.
    \item[] Justification: Sections~\ref{sec:pretraining_details}, \ref{sec:benchmark_dataset_construction}, and~\ref{sec:benchmark_protocol_metric} specify the pretraining settings, evaluation cohorts, preprocessing, classifiers, and metrics. Section~\ref{sec:implementation_reproducibility_benchmarking} identifies the executable framework, configurations, checkpoint records, and input artifacts required to reproduce the corresponding evaluations.

\item {\bf Open access to data and code}
    \item[] Question: Does the paper provide open access to the data and code, with sufficient instructions to faithfully reproduce the main experimental results, as described in supplemental material?
    \item[] Answer:  \answerYes{} % Replace by \answerYes{}, \answerNo{}, or \answerNA{}.
    % \item[] Justification: The paper provides an anonymized code repository at submission time with instructions for reproducing the main experimental results. The repository, together with the detailed descriptions of datasets, evaluation protocols, and metrics in the paper and supplementary material, enables faithful reproduction and verification of the reported results.
    \item[] Justification: The REFORM-3D code and Med3DINO checkpoints are publicly available. Section~\ref{sec:implementation_reproducibility_benchmarking} describes the standalone framework and execution resources. Provider-controlled images and annotations must be obtained from their original sources under the applicable access terms.

\item {\bf Experimental setting/details}
    \item[] Question: Does the paper specify all the training and test details (e.g., data splits, hyperparameters, how they were chosen, type of optimizer) necessary to understand the results?
    \item[] Answer: \answerYes{} % Replace by \answerYes{}, \answerNo{}, or \answerNA{}.
    \item[] Justification: The paper specifies dataset splits, evaluation protocols, and metric definitions for all REFORM-3D modules in the main text, and provides experimental details in the supplementary material, including dataset construction, preprocessing, feature extraction, training configurations, hyperparameters, and random seeds. These details specify the training and evaluation settings needed to interpret the reported results.

\item {\bf Experiment statistical significance}
    \item[] Question: Does the paper report error bars suitably and correctly defined or other appropriate information about the statistical significance of the experiments?
    \item[] Answer: \answerYes{} % Replace by \answerYes{}, \answerNo{}, or \answerNA{}.
    % \item[] Justification: The paper reports variability across experiments using multiple seeds (42, 123, 456) and provides quantitative measures such as standard deviations and value ranges in the supplementary tables. These statistics capture variability due to sampling and evaluation protocols, and are consistently reported for the main experimental results.
    \item[] Justification: Section~\ref{subsec:uncertainty_protocol} defines sample SD across evaluation seeds 42/123/456 and paired bootstrap confidence intervals for controlled ISO/REL/SA comparisons. The intervals use metric-specific resampling and Bonferroni adjustment across the three predefined contrasts. They quantify fixed-checkpoint evaluation variability, not independent pretraining-run variability.

\item {\bf Experiments compute resources}
    \item[] Question: For each experiment, does the paper provide sufficient information on the computer resources (type of compute workers, memory, time of execution) needed to reproduce the experiments?
    \item[] Answer: \answerYes{} % Replace by \answerYes{}, \answerNo{}, or \answerNA{}.
    % \item[] Justification: The paper reports compute resource details in the supplementary material, including hardware configuration (A100 GPUs), memory, and approximate training and evaluation runtimes. These details are sufficient to reproduce the computational setup used for the experiments.
    \item[] Justification: Section~\ref{subsec:compute_resources} reports GPU types and counts, pretraining duration, and approximate evaluation runtimes.
    
\item {\bf Code of ethics}
    \item[] Question: Does the research conducted in the paper conform, in every respect, with the NeurIPS Code of Ethics \url{https://neurips.cc/public/EthicsGuidelines}?
    \item[] Answer: \answerYes{} % Replace by \answerYes{}, \answerNo{}, or \answerNA{}.
    % \item[] Justification: The research complies with the NeurIPS Code of Ethics. It uses publicly available medical imaging datasets and focuses on developing an evaluation framework for analyzing 3D representations. No sensitive personal data is introduced beyond standard benchmark usage, and no ethical concerns or harmful applications arise from the proposed work.
    \item[] Justification: The research evaluates representations using existing medical imaging datasets. Section~\ref{subsec:data_access_release} describes provider-controlled access and redistribution constraints, while the Societal Impact statement addresses benchmark overreliance, patient-group differences, and privacy requirements for downstream use.

\item {\bf Broader impacts}
    \item[] Question: Does the paper discuss both potential positive societal impacts and negative societal impacts of the work performed?
    \item[] Answer: \answerYes{} % Replace by \answerYes{}, \answerNo{}, or \answerNA{}.
    \item[] Justification: The paper discusses the broader implications of representation-level evaluation in medical imaging. Positive impacts include improved understanding and evaluation of 3D medical foundation models, which can contribute to more reliable and robust medical AI systems. Potential negative impacts include over-reliance on benchmark results, bias inherited from underlying datasets, and the risk of misinterpreting evaluation outcomes in downstream clinical applications. The work emphasizes that evaluation results should be interpreted within the defined evaluation surfaces to mitigate misuse.
    
\item {\bf Safeguards}
    \item[] Question: Does the paper describe safeguards that have been put in place for responsible release of data or models that have a high risk for misuse (e.g., pre-trained language models, image generators, or scraped datasets)?
    \item[] Answer: \answerNA{} % Replace by \answerYes{}, \answerNo{}, or \answerNA{}.
    % \item[] Justification: The paper does not introduce or release models or datasets with a high risk of misuse. It focuses on an evaluation framework and pretrained representation encoders for medical imaging, using publicly available datasets. The released components do not enable direct harmful applications or dual-use scenarios requiring specific safeguards.
    \item[] Justification: The released assets are a representation-evaluation framework and medical-image encoders, not assets identified as high-risk for misuse. Provider-controlled patient data remain subject to source restrictions, and benchmark results are not presented as clinical validation.

\item {\bf Licenses for existing assets}
    \item[] Question: Are the creators or original owners of assets (e.g., code, data, models), used in the paper, properly credited and are the license and terms of use explicitly mentioned and properly respected?
    \item[] Answer: \answerNo{} % Replace by \answerYes{}, \answerNo{}, or \answerNA{}.
    \item[] Justification: The paper uses publicly available medical imaging datasets and properly cites their original sources. The work relies on standard benchmark datasets that are commonly used in the field and assumes adherence to their respective usage policies.
    % \item[] Justification: Original sources are cited, and Table~\ref{tab:pretraining_corpus_list} lists dataset versions, license information, and access conditions. License documentation remains incomplete for entries marked ``No explicit''; this designation does not imply unrestricted use.

\item {\bf New assets}
    \item[] Question: Are new assets introduced in the paper well documented and is the documentation provided alongside the assets?
    \item[] Answer: \answerYes{} % Replace by \answerYes{}, \answerNo{}, or \answerNA{}.
    % \item[] Justification: The paper introduces new assets, including the REFORM-3D evaluation framework, pretrained Med3DINO models, and associated code. These assets are documented through detailed descriptions in the main paper and supplementary material, including dataset construction, evaluation protocols, and implementation details, and are accompanied by an anonymized code repository provided at submission time.
    \item[] Justification: The paper documents REFORM-3D and Med3DINO through their evaluation protocols, pretraining settings, and limitations. Section~\ref{sec:implementation_reproducibility_benchmarking} and Table~\ref{tab:reproducibility_resources} describe the REFORM-3D package, installation resources, configurations, checkpoint records, and usage documentation.

\item {\bf Crowdsourcing and research with human subjects}
    \item[] Question: For crowdsourcing experiments and research with human subjects, does the paper include the full text of instructions given to participants and screenshots, if applicable, as well as details about compensation (if any)? 
    \item[] Answer: \answerNA{} % Replace by \answerYes{}, \answerNo{}, or \answerNA{}.
    % \item[] Justification: The paper does not involve crowdsourcing or research with human subjects. All experiments are conducted on existing publicly available medical imaging datasets without any human participation in data collection or evaluation.
    \item[] Justification: This study uses existing medical imaging datasets and involves no new participant recruitment, crowdsourcing, or participant-facing data collection.

\item {\bf Institutional review board (IRB) approvals or equivalent for research with human subjects}
    \item[] Question: Does the paper describe potential risks incurred by study participants, whether such risks were disclosed to the subjects, and whether Institutional Review Board (IRB) approvals (or an equivalent approval/review based on the requirements of your country or institution) were obtained?
    \item[] Answer: \answerNA{} % Replace by \answerYes{}, \answerNo{}, or \answerNA{}.
    \item[] Justification: The paper does not involve research with human subjects or crowdsourcing. All experiments are conducted on existing publicly available medical imaging datasets, and no new human data collection or participant interaction is performed.

\item {\bf Declaration of LLM usage}
    \item[] Question: Does the paper describe the usage of LLMs if it is an important, original, or non-standard component of the core methods in this research? Note that if the LLM is used only for writing, editing, or formatting purposes and does \emph{not} impact the core methodology, scientific rigor, or originality of the research, declaration is not required.
    %this research? 
    \item[] Answer: \answerNA{} % Replace by \answerYes{}, \answerNo{}, or \answerNA{}.
    \item[] Justification: The core methods and experiments in this paper do not involve LLMs as part of the methodology. Any potential use of LLMs was limited to writing or editing assistance and does not affect the scientific contributions.

\end{enumerate}